\documentclass{wscpaperproc}
\usepackage{latexsym}
\usepackage{graphicx}
\usepackage{mathptmx}
\usepackage[T1]{fontenc}

\usepackage{amsmath}
\usepackage{amsfonts}
\usepackage{amssymb}
\usepackage{amsbsy}
\usepackage{amsthm}

\usepackage{subfigure}
\usepackage{algpseudocode}
\usepackage{algorithm}
\usepackage{algcompatible}
\usepackage{multicol}
\usepackage{wrapfig}
\usepackage{lipsum}
\usepackage{xcolor}
\usepackage{booktabs,adjustbox}

\DeclareMathOperator*{\argmax}{arg\,max}
\DeclareMathOperator*{\argmin}{arg\,min}

\newcommand{\new}[1]{\textcolor{black}{ #1}}

\usepackage[pdftex,colorlinks=true,urlcolor=blue,citecolor=black,anchorcolor=black,linkcolor=black]{hyperref}

\newtheoremstyle{wsc}% hnamei
{3pt}% hSpace abovei
{3pt}% hSpace belowi
{}% hBody fonti
{}% hIndent amounti1
{\bf}% hTheorem head fontbf
{}% hPunctuation after theorem headi
{.5em}% hSpace after theorem headi2
{}% hTheorem head spec (can be left empty, meaning `normal')i

\theoremstyle{wsc}

\begin{document}

%***************************************************************************
% AUTHOR: AUTHOR NAMES GO HERE
% FORMAT AUTHORS NAMES Like: Author1, Author2 and Author3 (last names)
%
%		You need to change the author listing below!
%               Please list ALL authors using last name only, separate by a comma except
%               for the last author, separate with "and"
%

% setting up general page style
\pagestyle{fancyplain}

% setting up page style of first page
\thispagestyle{plain}
\firstPageHead{}

% setting up running header (authors) of subsequent pages
\chead{\fancyplain{}{\itshape Nezami, and Anahideh}}

% setting up seperation parameters
%\headsep=72pt
\rhead{}
\cfoot{}
\renewcommand{\headrulewidth}{0pt} % (renewcommand needed in fancyhdr to remove top decorative line)
%\headrulewidth=0pt  % ("setlength" needed in fancyheading to remove top decorative line)

%%%%%%%%%%%%%%%%%%%%%%%%%%%%%%%%%%%%%%%%%%%%%%%%%%%%%%%%%%%%%%%%%%%%%%%%%%%%%%
%                                                                            %
%     THESE COMMANDS ARE REQUIRED TO WORK WITH WSC.BST TO MAKE BIBLIO     %
%                                                                            %
%%%%%%%%%%%%%%%%%%%%%%%%%%%%%%%%%%%%%%%%%%%%%%%%%%%%%%%%%%%%%%%%%%%%%%%%%%%%%%
\makeatletter
\let\@internalcite\cite
\def\cite{\def\@citeseppen{-1000}%
    \def\@cite##1##2{(##1\if@tempswa , ##2\fi)}%
    \def\citeauthoryear##1##2##3{##1 ##3}\@internalcite}
\def\citeNP{\def\@citeseppen{-1000}%
    \def\@cite##1##2{##1\if@tempswa , ##2\fi}%
    \def\citeauthoryear##1##2##3{##1 ##3}\@internalcite}
\def\citeN{\def\@citeseppen{-1000}%
%  Pierre L'Ecuyer's fix for multiple cite bug
%  Added by Paul J Sanchez on 4 October 2001
%   \def\@cite##1##2{##1\if@tempswa , ##2)\else{)}\fi}%
%   \def\citeauthoryear##1##2##3{##1 (##3}\@citedata}
    \def\@cite##1##2{##1\if@tempswa, ##2)\else{}\fi}%
    \def\citeauthoryear##1##2##3{##1 (##3)}\@citedata}
\def\citeA{\def\@citeseppen{-1000}%
    \def\@cite##1##2{(##1\if@tempswa , ##2\fi)}%
    \def\citeauthoryear##1##2##3{##1}\@internalcite}
\def\citeANP{\def\@citeseppen{-1000}%
    \def\@cite##1##2{##1\if@tempswa , ##2\fi}%
    \def\citeauthoryear##1##2##3{##1}\@internalcite}
\def\shortcite{\def\@citeseppen{-1000}%
    \def\@cite##1##2{(##1\if@tempswa , ##2\fi)}%
    \def\citeauthoryear##1##2##3{##2 ##3}\@internalcite}
\def\shortciteNP{\def\@citeseppen{-1000}%
    \def\@cite##1##2{##1\if@tempswa , ##2\fi}%
    \def\citeauthoryear##1##2##3{##2 ##3}\@internalcite}
\def\shortciteN{\def\@citeseppen{-1000}%
%  Pierre L'Ecuyer's fix for multiple cite bug
%  Added by Paul J Sanchez on 2 September 2002
%  should have caught this last year...
%   \def\@cite##1##2{##1\if@tempswa , ##2)\else{)}\fi}%
%   \def\citeauthoryear##1##2##3{##2 (##3}\@citedata}
% Shane G. Henderson fix for extra right bracket at end of optional material June 8, 2005
%    \def\@cite##1##2{##1\if@tempswa, ##2)\else{}\fi}%
    \def\@cite##1##2{##1\if@tempswa, ##2\else{}\fi}%
    \def\citeauthoryear##1##2##3{##2 (##3)}\@citedata}
\def\shortciteA{\def\@citeseppen{-1000}%
    \def\@cite##1##2{(##1\if@tempswa , ##2\fi)}%
    \def\citeauthoryear##1##2##3{##2}\@internalcite}
\def\shortciteANP{\def\@citeseppen{-1000}%
    \def\@cite##1##2{##1\if@tempswa , ##2\fi}%
    \def\citeauthoryear##1##2##3{##2}\@internalcite}
\def\citeyear{\def\@citeseppen{-1000}%
    \def\@cite##1##2{(##1\if@tempswa , ##2\fi)}%
    \def\citeauthoryear##1##2##3{##3}\@citedata}
\def\citeyearNP{\def\@citeseppen{-1000}%
    \def\@cite##1##2{##1\if@tempswa , ##2\fi}%
    \def\citeauthoryear##1##2##3{##3}\@citedata}
%
% \@citedata and \@citedatax:
%
% Place commas in-between citations in the same \citeyear, \citeyearNP,
% \citeN, or \shortciteN command.
% Use something like \citeN{ref1,ref2,ref3} and \citeN{ref4} for a list.
%
\def\@citedata{%
    \@ifnextchar [{\@tempswatrue\@citedatax}%
                  {\@tempswafalse\@citedatax[]}%
}

\def\@citedatax[#1]#2{%
\if@filesw\immediate\write\@auxout{\string\citation{#2}}\fi%
  \def\@citea{}\@cite{\@for\@citeb:=#2\do%
    {\@citea\def\@citea{, }\@ifundefined% by Young
       {b@\@citeb}{{\bf ?}%
       \@warning{Citation `\@citeb' on page \thepage \space undefined}}%
{\csname b@\@citeb\endcsname}}}{#1}}%

% don't box citations, separate with ; and a space
% also, make the penalty between citations negative: a good place to break.
%
\def\@citex[#1]#2{%
\if@filesw\immediate\write\@auxout{\string\citation{#2}}\fi%
  \def\@citea{}\@cite{\@for\@citeb:=#2\do%
    {\@citea\def\@citea{; }\@ifundefined% by Young
       {b@\@citeb}{{\bf ?}%
       \@warning{Citation `\@citeb' on page \thepage \space undefined}}%
{\csname b@\@citeb\endcsname}}}{#1}}%

% (from apalike.sty)
% No labels in the bibliography.
%
\def\@biblabel#1{}
\makeatother

%\newlength{\bibhang}
%\setlength{\bibhang}{2em}

% Indent second and subsequent lines of bibliographic entries. Taken
% from openbib.sty: \newblock is set to {}.
% \renewcommand{\refname}{REFERENCES}

\newdimen\bibindent
\bibindent=0.0em
% SEC: was \def\thebibliography#1{\section*{\refname\@mkboth
% SEC: was   {\uppercase{\refname}}{\uppercase{\refname}}}\list
\def\thebibliography#1{\section*{\refname}\list
   {}{\settowidth\labelwidth{[#1]}
   \leftmargin\parindent
   \itemindent -\parindent
   \listparindent \itemindent
   \itemsep 0pt
   \parsep 0pt}
   \def\newblock{}
   \sloppy
   \sfcode`\.=1000\relax}

           % Set up BiBTeX macros

% needed to make the tex document look more like the word counterpart :-(
\setlength{\baselineskip}{12.7pt}

% AUTHOR: Enter the title, all letters in upper case
% \title{OPTIMIZING THE OPTIMIZERS: AN AUTOMATED HYPERPARAMETER ADAPTIVE SEARCH FOR SURROGATE OPTIMIZATION}

\title{
Hyperparameter Adaptive Search for Surrogate Optimization: A Self-Adjusting Approach
}

\author{Nazanin Nezami\\
    Hadis Anahideh\\
    \\
	Department of Mechanical and Industrial Engineering\\
	University of Illinois at Chicago\\
	842 W Taylor St\\
	Chicago, IL 60607, USA\\}
 
% \author{Nazanin Nezami\\[12pt]
%     Hadis Anahideh\\[12pt]
% 	Department of Mechanical and Industrial Engineering\\
% 	University of Illinois at Chicago\\
% 	842 W Taylor St\\
% 	Chicago, IL 60607, USA\\}

\maketitle

% These algorithms rely on multiple hyperparameters associated with sampling and surrogate fitting steps, which significantly affect their performance across various problems. In this study, we first investigate the impact of these hyperparameters on the performance of various SO algorithms.
% To automate the choice of hyperparameters, we propose a novel approach that adaptively modifies the hyperparameters' values based on the algorithm outcome. This approach aims to make SO algorithms more accessible to practitioners and ensures their effectiveness and faster convergence. Our approach enables us to both determine and modify the most influential hyperparameters for a given problem and SO approach, thus reducing the need for manual tuning. 

%\new{ABSTRACT should not exceed 150 words, no keywords}
\section*{ABSTRACT}
\new{
Surrogate Optimization (SO) algorithms have shown promise for optimizing expensive black-box functions. However, their performance is heavily influenced by hyperparameters related to sampling and surrogate fitting, which poses a challenge to their widespread adoption. We investigate the impact of hyperparameters on various SO algorithms and propose a Hyperparameter Adaptive Search for SO (HASSO) approach. HASSO is not a hyperparameter tuning algorithm, but a generic self-adjusting SO algorithm that dynamically tunes its own hyperparameters while concurrently optimizing the primary objective function, without requiring additional evaluations. The aim is to improve the accessibility, effectiveness, and convergence speed of SO algorithms for practitioners.
Our approach identifies and modifies the most influential hyperparameters specific to each problem and SO approach, reducing the need for manual tuning without significantly increasing the computational burden.
Experimental results demonstrate the effectiveness of HASSO in enhancing the performance of various SO algorithms across different global optimization test problems.
}

\section{INTRODUCTION}\label{sec:intro}

Optimizing expensive black-box functions is a crucial and challenging task that arises in various real-world applications, including engineering design \shortcite{chen2019aerodynamic}, financial management \shortcite{jafar2022financial}, drug discovery \shortcite{pyzer2018bayesian}, and complex algorithms \shortcite{feurer2019hyperparameter}. In such cases, evaluating the objective function is computationally expensive, and therefore, the evaluation budget is restricted. One of the most promising approaches to overcoming this challenge is to use SO algorithms \shortcite{anahideh2022high}. These algorithms involve constructing a cheap-to-evaluate model, known as a surrogate, that approximates the behavior of the black-box function. The surrogate model is then used to identify promising regions in the input space for further sampling and exploration.

Despite their effectiveness, SO algorithms heavily depend on multiple hyperparameters associated with sampling and surrogate fitting steps, leading to a significant impact on their performance across different problems. The selection of hyperparameters depends on several factors, such as the problem's characteristics, the complexity of the SO algorithm, and the available computational budget. The traditional approach for hyperparameter tuning involves manually adjusting hyperparameters using trial and error or grid search, either before or periodically during the optimization process. However, this approach can be tedious and time-consuming, especially when the number of hyperparameters and their search space is large. 
\new{Hence, a self-adaptive SO approach that simultaneously tunes hyperparameters while optimizing the primary objective offers a promising solution to reduce the time and effort required to find global optima.}
% the optimal hyperparameters, allowing for more efficient and effective surrogate optimization.

To overcome this challenge, a few studies in the field \new{integrates the hyperparameter tuning process within the optimization framework, this approach automates the selection of hyperparameters and adapts them dynamically during the optimization process.} 
% is a promising approach for improving the performance of surrogate optimization algorithms. 
\citeN{malkomes2018automating} proposed an automated Bayesian optimization approach that dynamically selects promising Gaussian Process (GP) models from a set of candidate models based on the model evidence metric (marginal likelihood), which measures how well the model explains the observed data.
Likewise, \shortciteN{benjamins2022towards} demonstrated the benefits of dynamically selecting acquisition functions for Bayesian optimization design by training an acquisition function selector. The selector predicts which acquisition function will have the best performance and switches to that function.
\new{Existing research in SO has predominantly focused on tuning GP hyperparameters, overlooking hyperparameters in candidate generation and sampling steps. Additionally, most research has concentrated on individual hyperparameters, neglecting the joint tuning of multiple hyperparameters. To address these limitations, a more flexible approach is required that can handle diverse surrogate models and all other relevant hyperparameters. By broadening the scope of hyperparameter tuning and considering their collective adjustment within the SO framework, without the need for additional evaluations, we can significantly enhance the performance and versatility of SO algorithms.}
% Although current research has studied the effect of Gaussian Process hyperparameters on the performance of SO algorithms, there is a need for a more flexible approach that can be applied to a broader range of surrogate models. Furthermore, the impact of hyperparameters in candidate generation and sampling steps of SO have not been explored in the existing literature. 
% Additionally, most of the existing research has focused on the impact of individual hyperparameters, and dynamically adjusting multiple hyperparameters has not been well studied.

\new{
\emph{Hyperparameter Adaptive Search for Surrogate Optimization (HASSO)} is a novel approach that automates the selection of hyperparameters for SO algorithms while aiming for the primary global optima. HASSO incorporates a secondary search on the hyperparameter space alongside the primary search on the decision variables of the optimization problem. Drawing inspiration from Thompson sampling \shortcite{agrawal2012analysis}, HASSO dynamically tunes its own hyperparameters within each SO iteration. It treats each hyperparameter as an arm with a predefined range of values and adapts them based on the probability of improving the primary objective value. HASSO achieves self-adjustment without exhaustively searching for hyperparameters at every step, resulting in significant computational resources and time savings. Unlike existing methods \shortcite{malkomes2018automating}, HASSO is not limited to surrogate model hyperparameters, making it versatile and adaptable to any SO algorithm. Experimental results validate the superior optimization performance of HASSO compared to several baselines, including fixed configurations, random updates, predefined rule-based updates, and grid search for hierarchical selection. HASSO outperforms these approaches across a diverse range of test problems, demonstrating its effectiveness and efficiency. By dynamically adapting hyperparameters within each iteration, HASSO achieves better convergence and exploration-exploitation balance, resulting in improved overall optimization performance.
}

%\vspace{-3mm}
The remainder of this paper is structured as follows: in Section~\ref{sec:background}, we provide background information on SO and discuss commonly used strategies. In Section~\ref{sec:impact}, we demonstrate the impact of hyperparameters on the performance of various SO algorithms. Section~\ref{sec:method} presents our proposed approach, and in Section~\ref{sec:result}, we report our experimental results. We discuss the implications of our findings and conclude in Section~\ref{sec:conclusion}.

\section{BACKGROUND}\label{sec:background}

\subsection{Surrogate Optimization}

% \nazanin{I integrated what we had in the algorithm here:}
% Surrogate optimization (SO) is an iterative sequential sampling procedure to optimize a black-box function $f$ over the closed hypercube $[\mathbf{x}^L, \mathbf{x}^U] \in \mathbb{R}^d$, where $\mathbf{x}$ is a $d$-dimensional input vector of variables, $\mathbf{x}^L$ and $\mathbf{x}^U$ are its coordinate-wise lower and upper bounds. 
\new{
Surrogate optimization (SO) is an iterative method to optimize a black-box function $f$ over a closed hypercube in $\mathbb{R}^d$. 
The hypercube is formed by the Cartesian product of $[x^{L}_1, x^{U}_1] \times [x^{L}_2, x^{U}_2] \times \dots \times [x^{L}_d, x^{U}_d]$, where $x^{L}_i$ and $x^{U}_i$ are the lower and upper bounds for the $i-$th dimension of an input vector $\mathbf{x}$.
%The hypercube is formed by the Cartesian product of $[x^{L}_1, x^{U}_1] \times [x^{L}_2, x^{U}_2] \times \dots \times [x^{L}_d, x^{U}_d] \subseteq$ , where $x^{L}_i$ and $x^{U}_i$ are the lower and upper bounds for the $i-$th dimension of an input vector $\mathbf{x}$. 
}
SO starts with generating a set of initial points, $\mathcal{D}=\{\mathbf{x}^1, \ldots, \mathbf{x}^{n}\}$, using a Design of Experiments (DOE) method, and record their corresponding function values in $\mathcal{F}= \{f(\mathbf{x}^i)|\mathbf{x}^i\in \mathcal{D}\}$.
Subsequently, a cheap-to-evaluate approximation model, $\hat{f}$, is fitted to the evaluated data points, $(\mathcal{D},\mathcal{F})$. A sample generation scheme, such as random sampling, is used to discretize the solution space and generate the candidate set $\mathcal{C}$. Next, an acquisition function is utilized to select candidate points with optimum acquisition values, $\mathcal{S}$, to be evaluated using the expensive black-box system. That is, $\mathcal{F}_ \mathcal{S}=\{f(\mathbf{x}^i)|\mathbf{x}^i\in \mathcal{S}\}$. 
Note that $\vert \mathcal{S} \vert=1$ for sequential, and $\vert \mathcal{S} \vert > 1$ for batch sampling algorithms. In each iteration, the newly evaluated points are appended to $\mathcal{D}$, $\mathcal{D}= \mathcal{D}\cup \mathcal{S}$; and their corresponding function values are added to $\mathcal{F}$, $\mathcal{F}=\mathcal{F}\cup \mathcal{F}_ \mathcal{S}$. The algorithm iterates until a termination criterion (e.g., maximum evaluation budget) is satisfied, and returns the best observed solution,  $\mathbf{x}^* \in \argmin_{\mathbf{x}\in \mathcal{D}} f(\mathbf{x})$.

\subsection{Surrogate Modeling}

%To implement Algorithm~\ref{alg:SO},
To implement SO algorithms, an efficient surrogate model is needed to approximate the expensive black-box function. Several types of surrogate models have been proposed in the literature, including Gaussian Processes (GP), Multivariate Adaptive Regression Splines (MARS), and Radial Basis Function (RBF) \shortcite{anahideh2022high}. 
% These models have demonstrated effectiveness in approximating black-box functions in a cost-effective manner.
% To implement Algorithm~\ref{alg:SO}, it is essential to employ a cost-effective surrogate model to approximate the black-box function. Gaussian Processes (GP), Multivariate Adaptive Regression Splines (MARS), and Radial Basis Function (RBF) are among the commonly used surrogate models proposed in the literature that have demonstrated effectiveness for this purpose.
\textbf{Gaussian processes (GPs)} \shortcite{rasmussen2003gaussian} are widely used for modeling expensive black-box functions due to their flexible yet powerful nature.
%GPs define a prior probability distribution over the space of functions, which is updated to a posterior distribution based on observed data. 
%\hadis{well I revised but still it is unclear on prior and posterior, elaborate} 
% GPs not only provide an approximation of the black-box function but also estimate the uncertainty of the prediction at each given point. 
% Given a set of evaluated observations $\mathcal{D}$, a GP can be used to model the predicted mean $\hat{\mu}(\mathbf{x})$ and the degree of uncertainty $\hat{\sigma}(\mathbf{x})$ at any given point $\mathbf{x}$ in the input space. 
Let $f(\mathbf{x})$ be the output corresponding to the newly evaluated input vector $\mathbf{x}$, $X$ represents the matrix of already evaluated points in $\mathcal{D}$, and $\textbf{f}$ be the corresponding vector of function values. The estimated mean and variance for $\mathbf{x}$ can be calculated as $\hat{\mu}(\mathbf{x})=K(X,\mathbf{x} )\Sigma^{-1} \textbf{f}$ and $\hat{\sigma}(\mathbf{x})= K(\mathbf{x} ,\mathbf{x})-K(X,\mathbf{x})\Sigma^{-1}K(\mathbf{x},X)$,
% based on the updated GP model as follows:
% \begin{equation}
% \label{eq:GP-mean}
% \hat{\mu}(\mathbf{x})=K(X,\mathbf{x} )\Sigma^{-1} \textbf{f},
% \end{equation}
% \begin{equation}
% \label{eq:GP-cov}
% \hat{\sigma}(\mathbf{x})= K(\mathbf{x} ,\mathbf{x})-K(X,\mathbf{x})\Sigma^{-1}K(\mathbf{x},X),
% \end{equation}
where $\Sigma$ represents the covariance matrix and $K$ is a positive semidefinite similarity kernel. \new{
The general mathematical functional form for a similarity kernel is given by $K(\mathbf{x}, \mathbf{x}') = \theta_0 \cdot \exp\left(-\frac{1}{2}(\mathbf{x} - \mathbf{x}')^T\boldsymbol{\Theta}^{-1}(\mathbf{x} - \mathbf{x}')\right)$.
% Here, $K(\mathbf{x}, \mathbf{x}')$ represents the similarity kernel, which measures the similarity between input points $\mathbf{x}$ and $\mathbf{x}'$. 
The kernel function is scaled by the hyperparameter $\theta_0$ and is parameterized by a matrix of hyperparameters $\boldsymbol{\Theta}$. The specific form of the matrix $\boldsymbol{\Theta}$ depends on the choice of the kernel and may include parameters such as lengthscale and amplitude associated with the chosen kernel.
The hyperparameters, denoted as $\boldsymbol{\Theta}$ and $\theta_0$, control the behavior of the similarity kernel and need to be tuned.}
Common kernel functions are Squared Exponential (SE), Matérn, the Periodic kernel, and Linear. Accurate tuning of the kernel and lengthscale is essential for effective GP modeling. 
% The off-diagonal elements of $K$, $K(\mathbf{x_i},\mathbf{x_j})$, are associated with any distinct given pair of points $\mathbf{x_i},\mathbf{x_j} \in \mathcal{X}$ that represent the correlation between data points.

% \nazanin{The lengthscale parameter is a part of this covariance function and controls the extent to which nearby points influence each other. A larger lengthscale implies a smoother while a smaller lengthscale indicates a more rapidly varying function.}

% Common kernel functions are Squared Exponential (SE) for smooth functions with long-range correlations, Matérn for varying levels of smoothness, the Periodic kernel for periodic behavior, and Linear for linear trends. Accurate tuning of the kernel and length scale is essential for effective GP modeling. 
% The length scale parameter in these kernels determines properties such as the range of correlation for SE/Matérn and the smoothness/periodicity for Periodic. Accurate tuning of the kernel and length scale is essential for effective GP modeling. 
% The lengthscale parameter in these kernels determines properties like range of correlation for SE/Matérn, and smoothness/periodicity for Periodic. Accurate tuning of kernel and lengthscale is crucial for effective GP modeling.
%By estimating the prediction errors, GPs can guide the sampling process to select points in promising regions with large variance, reducing the model uncertainty.

\subsection{Candidate Generation via Discretization }\label{sec:Discretization}
 
% Discretization is a crucial step in Black-Box Optimization (BBO), as it enables the use of acquisition functions that are optimized over a finite set of points or a grid, making the optimization more computationally efficient. % Discretizing the continuous solution space is a crucial step in the context of Black-Box Optimization (BBO). It is commonly used in Surrogate Optimization (SO) to address the computational challenges associated with evaluating acquisition functions. By discretizing the problem, we convert the continuous solution space into a discrete space, making the evaluation of acquisition functions more effective and tractable. % By discretizing the solution space, we can allocate evaluations to different regions, ensuring a thorough exploration of the search space while focusing on promising areas. 
\new{
Discretizing the continuous solution space is a crucial step in Black-Box Optimization (BBO) and SO. It addresses the computational challenges of evaluating acquisition functions over the continuous space by converting it into a discrete one. %This simplifies the evaluation process, improves computational efficiency, and enables a balanced exploration-exploitation trade-off.
Effective discretization reduces the complexity of the optimization problem by restricting the search space to a finite set of discrete points, facilitating the thorough exploration of the search space and focusing on promising areas. 
This helps in discovering the global optima efficiently. The choice of a robust discretization method is essential for effective optimization. Among the proposed techniques, dynamic coordinate search \shortcite{regis2013combining,nezami2022empirical} is the most promising approach. However, even in applying the dynamic coordinate search technique, there are hyperparameters to consider, such as the radius rule that determines the search neighborhood size. Selecting an appropriate radius rule is crucial for achieving a balance between exploration and exploitation, influencing optimization performance and convergence rate.
}

\subsection{Acquisition Functions for Sampling }\label{sec:Acquisition}
Acquisition functions play a critical role in guiding the search toward promising candidate points for sampling in SO. They balance exploration and exploitation by selecting points that either improve the model's understanding of unknown regions (exploration) or result in the best objective value (exploitation). Through an effective selection of candidate points, acquisition functions facilitate convergence toward the optimal solution. In the following, we will provide a brief overview of common acquisition functions as documented in the BBO literature.
% Acquisition functions are crucial in SO to select promising candidate points for sampling. They balance exploration and exploitation by choosing points that improve the model's understanding of unknown regions (exploration) or result in the best objective value (exploitation). By guiding the search towards promising points, acquisition functions enable efficient exploration of the solution space and convergence towards the optimal solution. We will now provide a concise overview of common acquisition functions as documented in the BBO literature. 

% \textbf{Probability of Improvement (PI)}~\shortcite{kushner1964new} calculates the probability that a new sample will improve the current best solution, $f^{o}_{n}$, and selects the point with the highest probability. Thus, PI tends to be more focused on exploitation. More formally, $PI(\mathbf{x})=P(I(\mathbf{x}))=P(f(\mathbf{x})>f^{o}_{n})= \Phi(\frac{\hat{\mu}(\mathbf{x})-f^{o}_{n}}{\hat{\sigma}(\mathbf{x})})$.

% , as it prioritizes points that are likely to have a higher objective value than the current best value.
% \vspace{-2mm}
% \begin{equation}
% PI(\mathbf{x})=P(I(\mathbf{x}))=P(f(\mathbf{x})>f^{o}_{n})= \Phi(\frac{\hat{\mu}(\mathbf{x})-f^{o}_{n}}{\hat{\sigma}(\mathbf{x})}),
% \vspace{-2mm}
% \end{equation} 

\textbf{Expected Improvement (EI)}~\shortciteN{jones1998efficient} selects the point with the highest expected improvement over the current best objective value. 
% It calculates the expected value of the difference between the current best objective value and the objective value of a new sample, weighted by its probability.
EI considers an explicit emphasis on both exploration and exploitation through the mean and the variance of the model's prediction.
% calculates the expected improvement of a new sample over the current best value, and select the points with highest expected improvement.
% EI can balance exploration-exploitation, as it considers both 
% through the mean and the variance of the model's prediction. 
% Let $f(\mathbf{x})|\mathcal{D}_n$ be normally distributed with mean $\hat{\mu}(\mathbf{x})$ and variance $\hat{\sigma}^2(\mathbf{x})$ for any given point $\mathbf{x}$. 
The closed-form equation of EI can be written as, $EI(\mathbf{x})=E(I(\mathbf{x}))= (\hat{\mu}(\mathbf{x})-f^{o}_{n}) \Phi(\frac{\hat{\mu}(\mathbf{x})-f^{o}_{n}}{\hat{\sigma}(\mathbf{x})})+\hat{\sigma}(\mathbf{x})\phi(\frac{\hat{\mu}(\mathbf{x})-f^{o}_{n}}{\hat{\sigma}(\mathbf{x})})$, 
% \vspace{-3mm}
% \begin{equation}
% \label{eq:EI}
% EI(\mathbf{x})=E(I(\mathbf{x}))= (\hat{\mu}(\mathbf{x})-f^{o}_{n}) \Phi(\frac{\hat{\mu}(\mathbf{x})-f^{o}_{n}}{\hat{\sigma}(\mathbf{x})})+\hat{\sigma}(\mathbf{x})\phi(\frac{\hat{\mu}(\mathbf{x})-f^{o}_{n}}{\hat{\sigma}(\mathbf{x})}),
% \vspace{-2mm}
% \end{equation}
where $I(\mathbf{x})=\max(0,[\hat{\mu}(\mathbf{x})-f^{o}_{n}]^{+})$, { and $f^{o}_{n}$ is best observed value after $n$ evaluations. The standard normal density and cumulative distribution are denoted by $\phi(\mathbf{x})$ and $\Phi(\mathbf{x})$, respectively.} 
% $\phi(\mathbf{x})$ denotes the  standard normal density function, and $\Phi(\mathbf{x})$ is standard normal cumulative distribution. 
% defines the EI improvement criteria at a given point $\mathbf{x}$. 

\textbf{Upper Confidence Bound (UCB)}~\shortciteN{srinivas2010gaussian} balances exploration and exploitation by selecting the point with the highest upper confidence bound. The upper confidence bound is calculated as the sum of the mean prediction ($\hat{\mu}(\mathbf{x})$) and a confidence interval term, which is determined by the variance of the model predictions ($\hat{\sigma}(\mathbf{x})$). 
% The UCB method tends to select points with high predicted means and high uncertainty. 
At any point $\mathbf{x}$, the UCB equation is given by $\alpha_{UCB}(\mathbf{x},\beta_t)=\hat{\mu}(\mathbf{x})+\beta_t^{0.5} \hat{\sigma}(\mathbf{x})$, where $\beta_t$ is a hyperparameter that controls the balance between exploration and exploitation at iteration $t$. 
% calculates an upper bound on the objective function value and selects the point with the highest value, which balances exploration and exploitation by considering both the mean ($\hat{\mu}(\mathbf{x})$) and the uncertainty of the model's prediction ($\hat{\sigma}(\mathbf{x})$). UCB tends to be more exploratory, as it prioritizes points that have higher uncertainty in their predictions. 
% The UCB equation at any point $\mathbf{x}$ could be described as, $\alpha_{UCB}(\mathbf{x},\beta_t)=\hat{\mu}(\mathbf{x})+\beta_t^{0.5} \hat{\sigma}(\mathbf{x}),$
% \vspace{-2mm}
% \begin{equation}
% \label{eq:UCB}
% \alpha_{UCB}(\mathbf{x},\beta_t)=\hat{\mu}(\mathbf{x})+\beta_t^{0.5} \hat{\sigma}(\mathbf{x}),
% \vspace{-2mm}
% \end{equation} 
% where $\beta_t$ serves as the trade-off hyperparameter. 
% Higher value of $\beta_t$ encourages more exploration while lower values promotes more exploitation. 

\textbf{Weighted Score (Wscore)}~\shortciteN{regis2013combining} is an acquisition function that balances exploration and exploitation by assigning scores to candidate points based on a linear aggregation over a distance metric and the estimated objective value obtained from the surrogate model. The acquisition function for Wscore at any candidate solution $\mathbf{x} \in \mathcal{C}$ is given by $W^{t}(\mathbf{x})=w_d^{t} \times V_d^{t}(\mathbf{x})+w^{t}_r \times V^{t}_r(\mathbf{x}),$
% \textbf{Weighted Score (Wscore)}~\shortcite{regis2013combining}
% assigns scores to candidate points based on a linear aggregation over a distance metric (exploration) and the estimated objective value obtained from the surrogate model (exploitation). The acquisition function for Wscore at any candidate solution $\mathbf{x} \in \mathcal{C}$ is given by: 
% \vspace{-2mm}
% \begin{equation}
% \label{eq:score-based}
% W^{t}(\mathbf{x})=w_d^{t} \times V_d^{t}(\mathbf{x})+w^{t}_r \times V^{t}_r(\mathbf{x}),
% \vspace{-2mm}
% \end{equation} 
where $w_d^{t}$ and $w_r^{t}$ are predefined weight patterns such that $w_d^{t}=1-w_r^{t}$, and $V_d^{t}(\mathbf{x})$ and $V_r^{t}(\mathbf{x})$ are functions that return normalized scores for the distance metric and the estimated objective value, respectively. The weights $w_d^{t}$ and $w_r^{t}$ determine the balance between exploration and exploitation in the acquisition function.

\subsection{Thompson Sampling} 
Thompson Sampling \shortcite{thompson1933likelihood} is a probabilistic decision-making strategy commonly used in both SO \shortcite{kandasamy2018parallelised} and Multi-Armed Bandit (MAB) \shortcite{agrawal2012analysis} problems. In SO, Thompson Sampling constructs a posterior distribution over the function based on the observed data and samples candidate points according to their probability of being the global optimum. Beta distributions are often used as the prior distribution for SO problems in Thompson Sampling. In MAB problems, Thompson Sampling constructs a posterior distribution over the reward distribution of each arm based on the observed rewards and selects the arm with the highest expected reward. The choice of prior distribution depends on the problem and the available prior knowledge. By balancing exploration and exploitation in a probabilistic way, Thompson Sampling can lead to faster convergence to the optimal solution.

\section{IMPACT OF ALGORITHMIC HYPERPARAMETERs ON SURROGATE OPTIMIZATION} \label{sec:impact}

\begin{figure*}[!thb]
\centering
\subfigure[Sampling trade-off: Ackley-5$d$]{\label{fig:imp1}\includegraphics[width=0.325\linewidth]{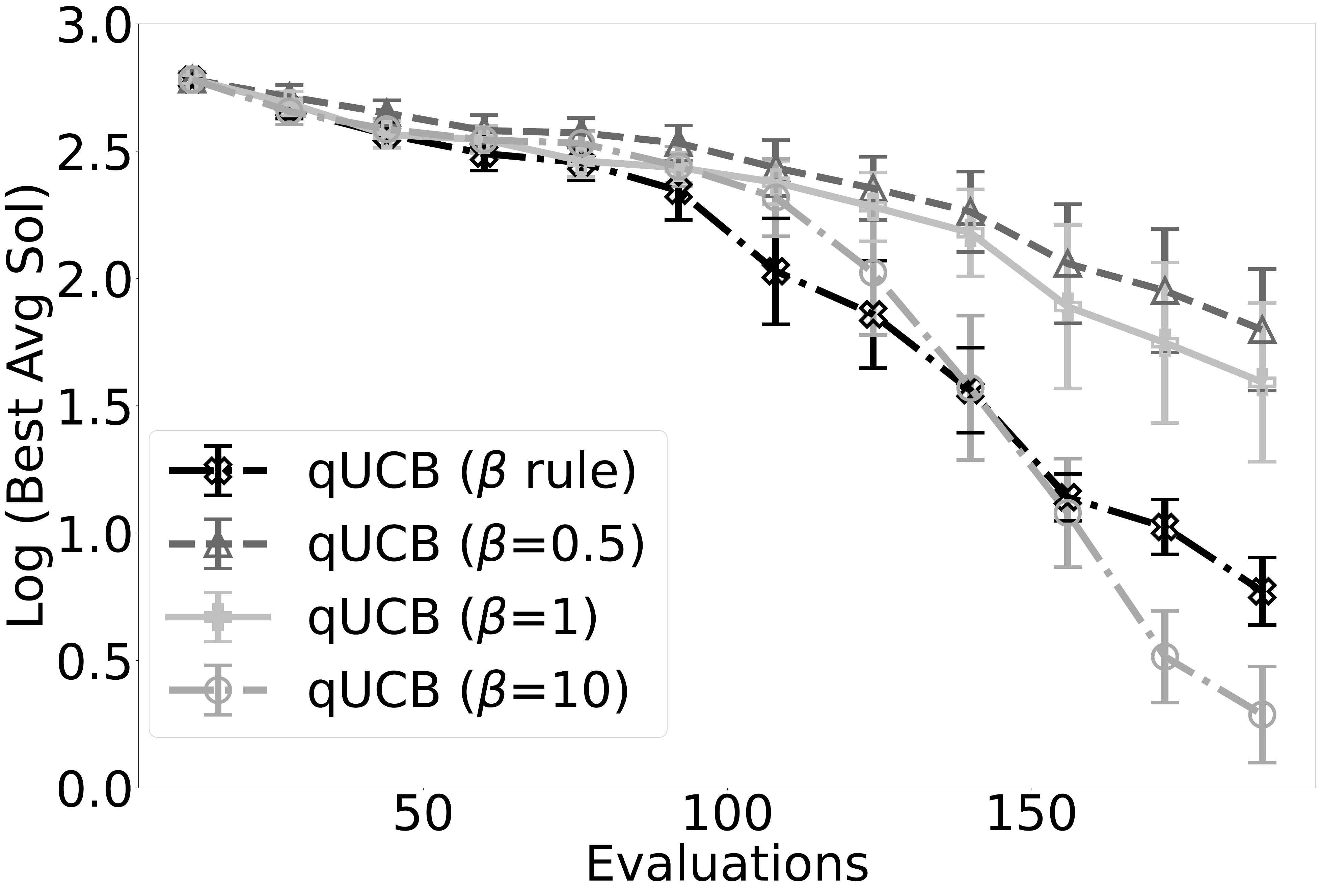}}
\subfigure[Sampling weight pattern: Ackley-5$d$]{\label{fig:imp2}\includegraphics[width=0.325\linewidth]{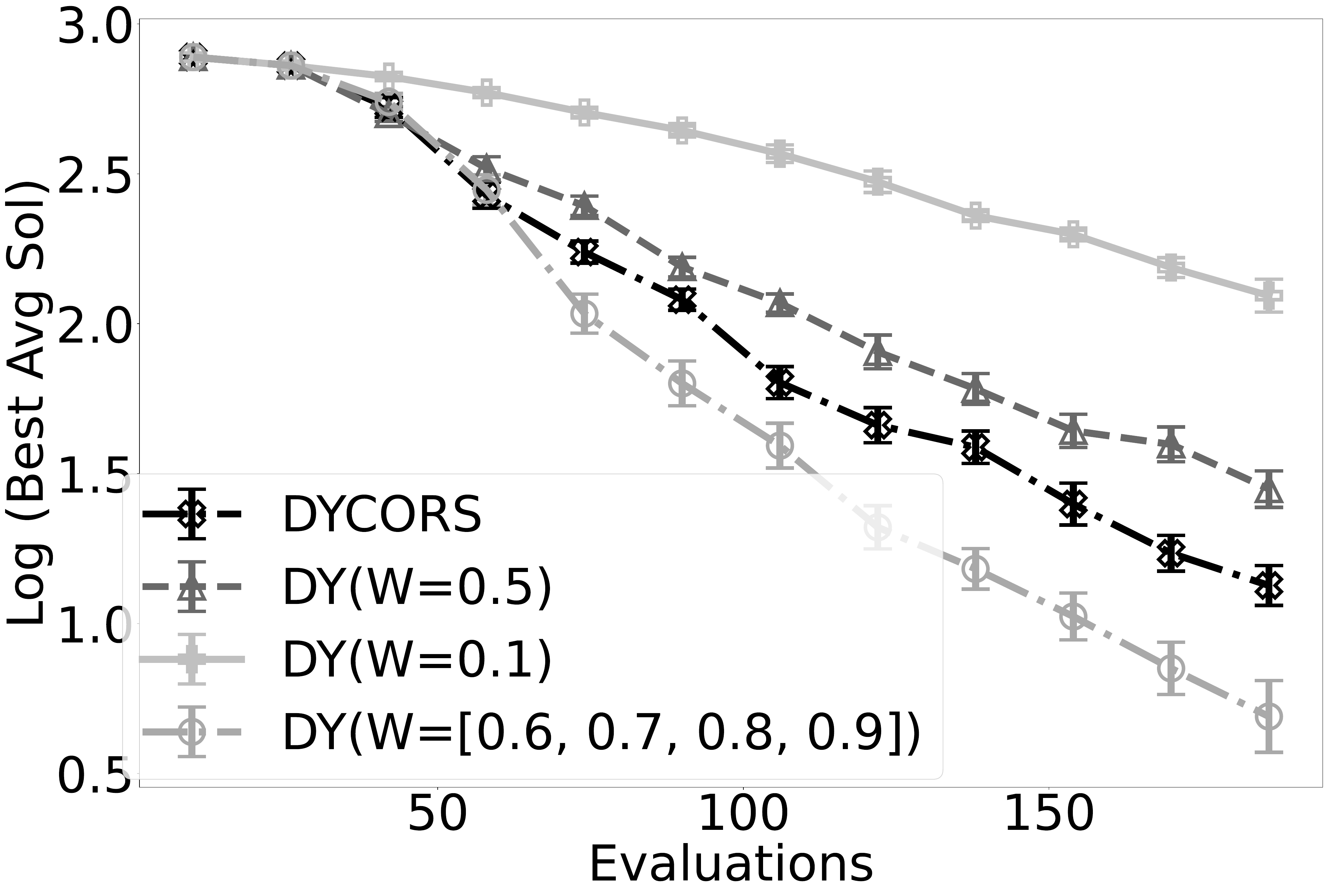}}
\subfigure[Discretization radius: Rosenbrock-15$d$]{\label{fig:imp3}\includegraphics[width=0.325\linewidth]{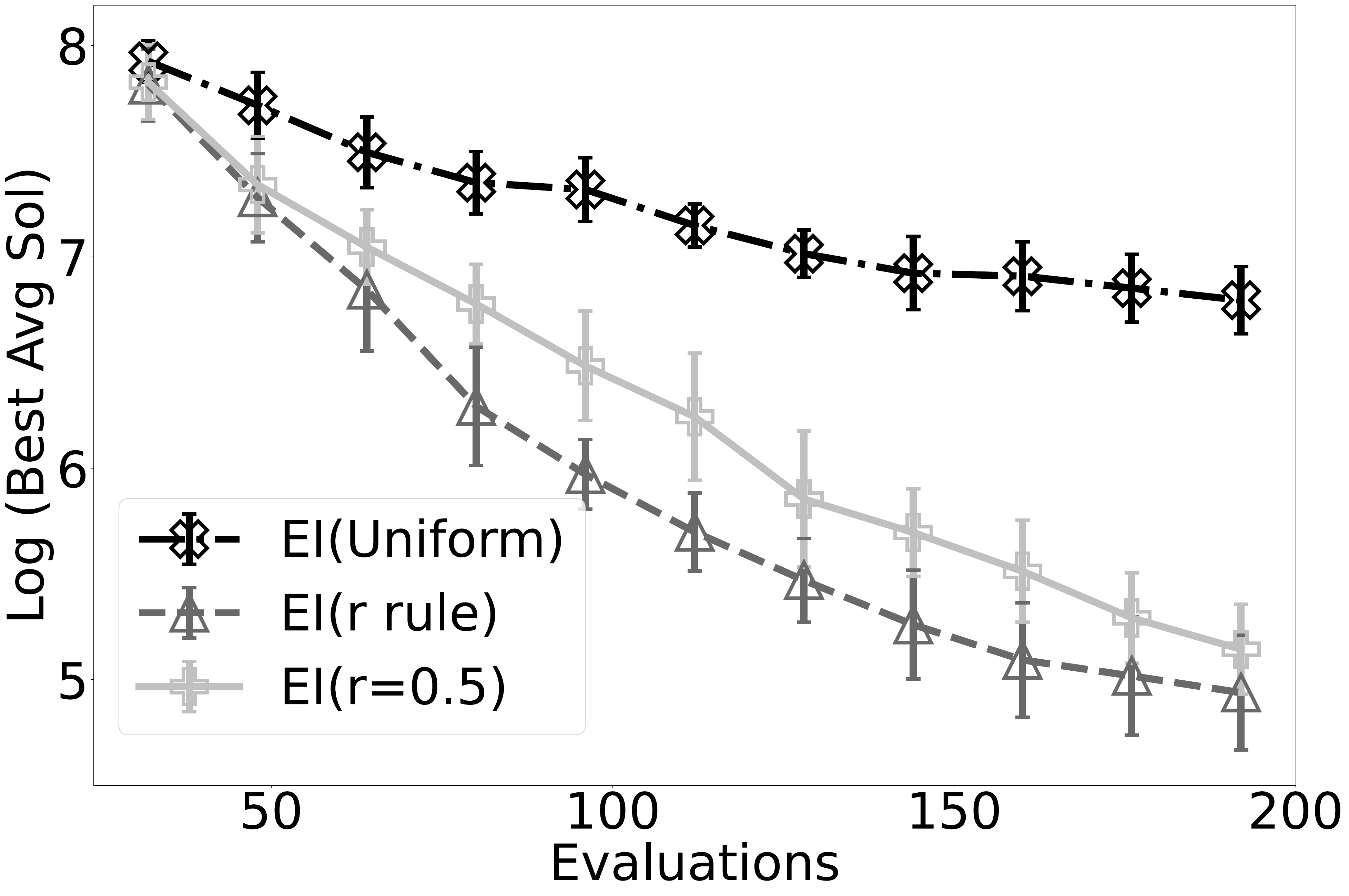}}

\subfigure[Discretization radius: DNA Binding]{\label{fig:imp4}\includegraphics[width=0.33\linewidth]{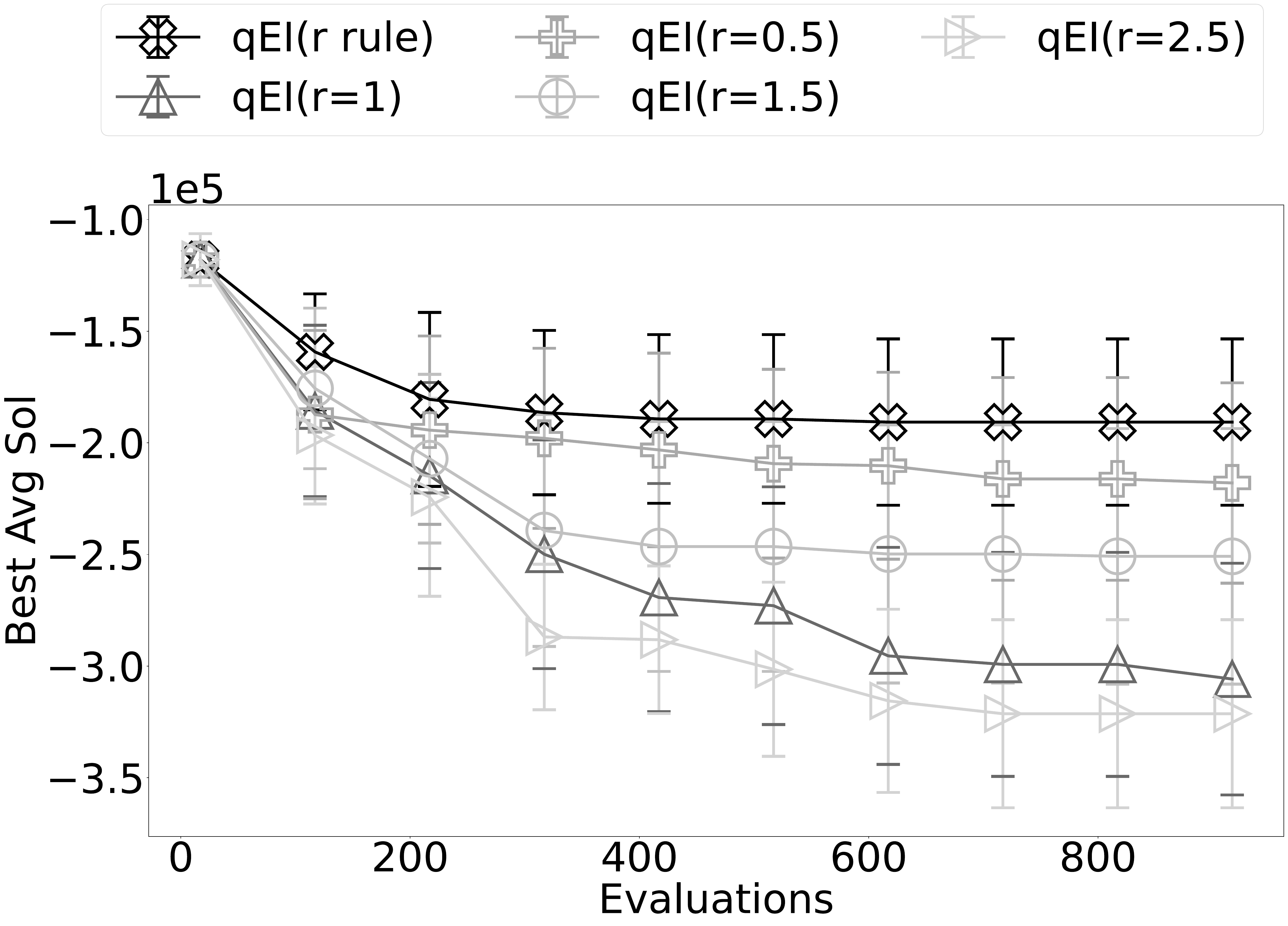}}
\subfigure[Radius=1: Ackley-5$d$]{\label{fig:imp5}\includegraphics[width=0.32\linewidth]{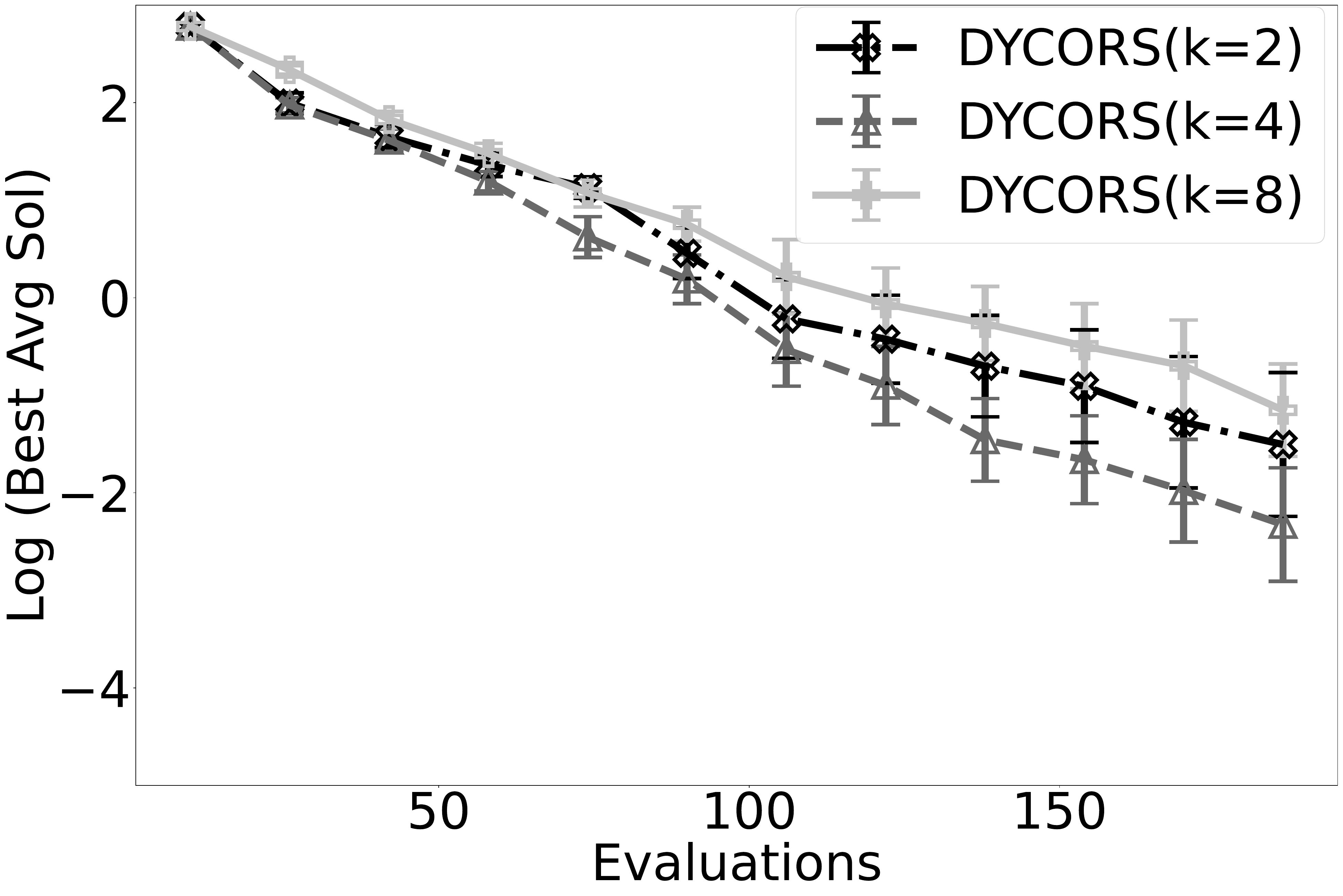}}
\subfigure[Radius with r-rule: Ackley-5$d$]{\label{fig:imp6}\includegraphics[width=0.32\linewidth]{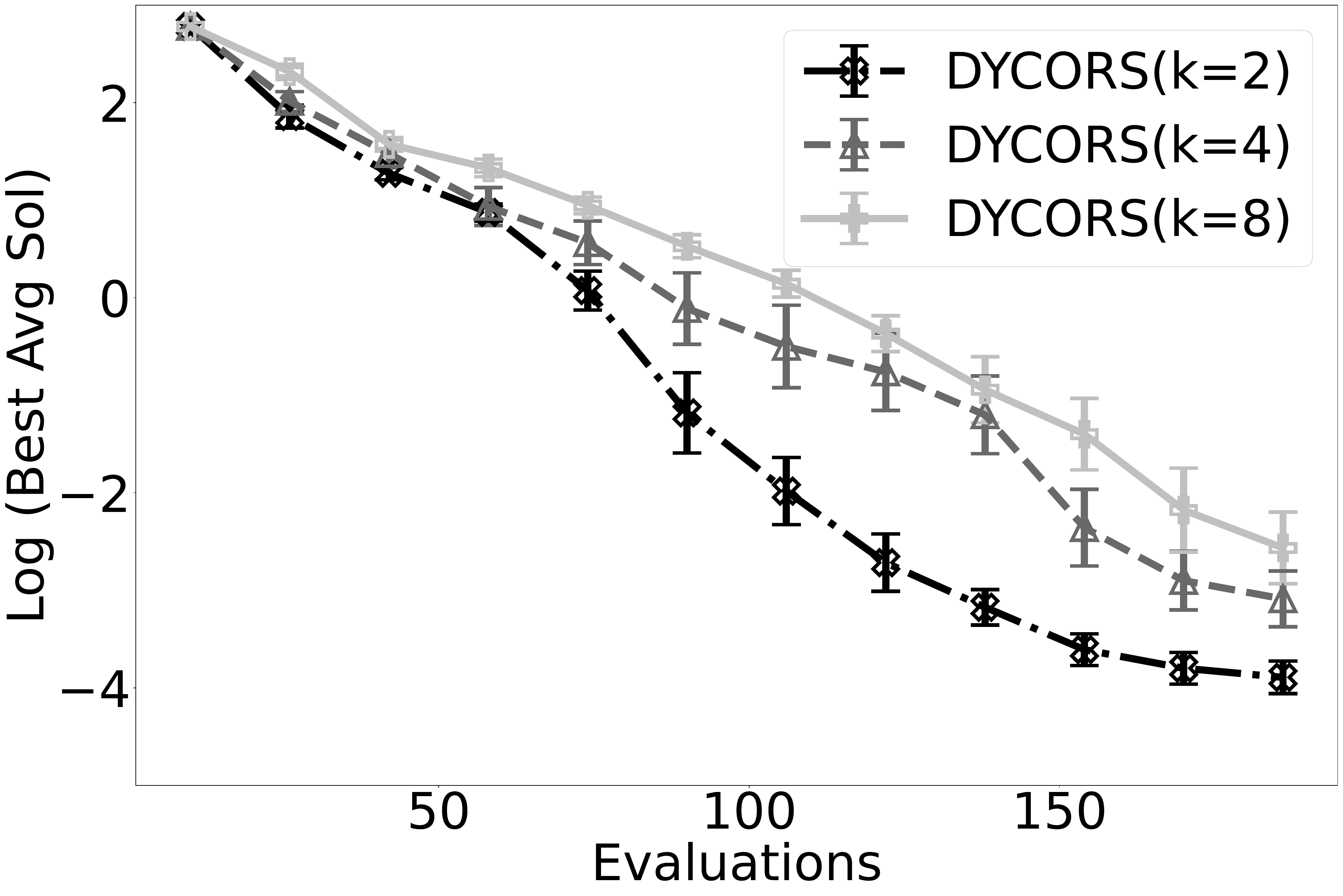}}

\caption{Performance of different SO algorithms under modified hyperparameters values.} 
\label{fig:imp}
\end{figure*}

As discussed in Section~\ref{sec:background}, the choice of hyperparameters can significantly affect the performance of SO algorithms. Hyperparameters are tunable parameters that exist in different stages of a given SO algorithm, such as modeling, discretization, and sampling. For instance, SO algorithms that adopt GPs probabilistic models as a surrogate often include hyperparameters such as the kernel function, lengthscale, and the prior probability distribution. The choice of surrogate hyperparameters can influence the convergence rate, the ability to find the global optimum, and the exploration-exploitation trade-off, which in turn affects the overall performance of the algorithm. A poor choice of hyperparameters can result in slow convergence or getting stuck in local optima, leading to suboptimal performance of the algorithm.
In this section, we elaborate on this phenomenon through illustrative examples shown in Figure~\ref{fig:imp}.  Specifically, we demonstrate the impact of hyperparameters on the Ackley global optimization test problem, utilizing the batch version of Upper Confidence Bound (qUCB), and Dynamic Coordinate Search (DYCORS) \shortcite{regis2013combining} in a 5-dimensional setting, shown in Figures~\ref{fig:imp1} and~\ref{fig:imp2}, respectively. 

Figure~\ref{fig:imp1} shows the impact of the trade-off control hyperparameter $\beta$ in the qUCB acquisition function, where different initial values for $\beta$ or a fixed rule to update it in each iteration significantly impact the qUCB convergence rate and the final solution obtained. Similarly, DYCORS, Figure~\ref{fig:imp2}, which uses the Wscore acquisition to assign scores to candidate points for sampling, is severely impacted by the weight pattern ($w_d^{t},w_r^{t}$) considered in each iteration. Furthermore, Figures~\ref{fig:imp3} and~\ref{fig:imp4} reveal the importance of the quality of the generated candidate points, through a discretization schema, for sampling with the EI acquisition function. Figure~\ref{fig:imp3} illustrates the impact of the dynamic coordinate-based discretization approach proposed in section~\ref{sec:Discretization} on the performance of the Expected Improvement (EI) acquisition function compared to the random uniform discretization schema for the 15-dimensional Rosenbrock global optimization test problem. The results demonstrate the disadvantage of using the random uniform discretization schema for EI. Additionally, our experiments on real-world DNA binding test problems, as shown in Figure~\ref{fig:imp4}, highlight the importance of the radius hyperparameter in the distribution of the generated candidate points resulting from a dynamic discretization strategy.
%The impact of this hyperparameter is shown in Figures~\ref{fig:imp3} and~\ref{fig:imp4} for the 15-dimensional Rosenbrock global optimization and the real-world DNA binding test problems, respectively. 
% \hadis{this needs to be added too: In addition, the discretization schema, random or coordination-based also matters} \hadis{the legend is once uniform and then rand, pick one.} \hadis{why do we have rand here in general? if we want to show the impact of r in DYCORS we should just show that.}
It is worth noting that this phenomenon becomes even more complex when employing SO to tackle real-world problems, particularly in the absence of domain-specific expert knowledge. {For instance, the relationship between the structure of DNA and its binding affinity with proteins is highly complex and cannot be easily modeled using a simple mathematical expression. Therefore, identifying the specific algorithmic criteria required to obtain an optimal solution for this problem is a significant challenge.} Hence, accurate tuning of algorithmic hyperparameters is crucial to ensure the effectiveness and efficiency of SO algorithms. %\hadis{what is the complexity observed for DNA? you should elaborate}

% The qUCB acquisition function in Figure~\ref{fig:imp1} includes a trade-off control hyperparameter $\beta$ in its acquisition function. Using different initial values for $\beta$ or fixing a rule to update it in each iteration, plays a significant role in the qUBC convergence rate and the final solution obtained. Similarly, DYCORS in Figure~\ref{fig:imp2} uses the Wscore acquisition to assign scores to candidate points for sampling. The performance of Wscore is majorly impacted by the considered weight pattern ($w_d^{t},w_r^{t}$) in each iteration. 
% Moreover, Figures~\ref{fig:imp3} and~\ref{fig:imp4} reveal the importance of the generated candidate points (discretization) for sampling with the EI acquisition function. Radius is a hyperparameter that plays a crucial role in the distribution of the generated candidate points resulting from a dynamic discretization strategy. Figures~\ref{fig:imp3} and~\ref{fig:imp4} show the significance of this impact for 15-dimensional Rosenbrock global optimization, and the real-world DNA binding test problems, respectively. 

% It is worth noting that the complexity of this phenomenon may be further exacerbated when employing SO to tackle real-world problems such as Figure~\ref{fig:imp4}, particularly in the absence of domain-specific expert knowledge.

While the impact of individual algorithmic hyperparameters cannot be disregarded, the potential interactions among different hyperparameters in various SO algorithms pose an additional challenge, as shown in Figures~\ref{fig:imp5} and \ref{fig:imp6}. In these examples, we use the DYCORS algorithm with the Ackley-5d test problem to illustrate the impact of changes in the radius hyperparameter for discretization and the batch size ($k$) for sampling.  
Specifically, we compare the results obtained with a fixed radius of $1$ in Figure~\ref{fig:imp5} to those obtained using the $r$-rule strategy in Figure~\ref{fig:imp6}, while varying the batch size. Our experimental results clearly show that the choice of both the batch size and radius hyperparameters has impacted the ability of the algorithm to find optimal solutions, and the impact of these hyperparameters cannot be considered in isolation. It is important to carefully consider the interplay between hyperparameters when designing and implementing SO algorithms. Hence, the complex interactions between hyperparameters can make the optimization process more challenging, necessitating careful experimentation and analysis to identify the best combination of hyperparameter settings. 
% Therefore, the intricate relationships between hyperparameters may further complicate the optimization process, requiring careful consideration and experimentation to identify the optimal combination of hyperparameter settings. 
This underscores the need for thorough sensitivity analysis and hyperparameter tuning in SO applications, to ensure reliable results.

\vspace{-3mm}

%\vspace{-4mm}
\section{HASSO: HYPERPARAMETER ADAPTIVE SEARCH FOR SURROGATE OPTIMIZATION}\label{sec:method}

\new{We present our self-adjusting method for adaptively tuning hyperparameters of SO while optimizing the primary objective. HASSO is versatile and can work with different SO algorithms, regardless of the surrogate model or acquisition/sampling function used. Although we focus on sequential BBO methods \shortcite{huang2006global,shahriari2015taking} in this study, our method can also be applied to parallel settings that involve batch evaluations \shortcite{krityakierne2016sop,wu2016parallel}.}

% In this section, we present our proposed approach for adaptive SO hyperparameter tuning, called \emph{HASSO}. Our approach can incorporate various SO algorithms and is agnostic to the surrogate model used and acquisition or sampling function. Although we focus on sequential BBO \shortcite{huang2006global,shahriari2015taking} approaches in this study, similar logic could be useful in parallel settings for batch evaluations \shortcite{krityakierne2016sop,wu2016parallel}. 
%Sequential BBO approaches are used to optimize a decision-making process or a system that evolves over time which must be solved sequentially. 
% \hadis{you do not need this last sentence but you can add citations for batch and sequential in the previous sentence}.

\new{Taking inspiration from the widely-used Thompson Sampling approach for solving the MAB problem \shortcite{agrawal2012analysis},  HASSO incorporates a hyperparameter tuning step within an SO algorithm. It treats each hyperparameter of a given SO algorithm as a separate arm with an unknown reward distribution.} 
HASSO utilizes the beta distribution which has two shape parameters, $\alpha$ and $\beta$, that can be interpreted as the number of successes and failures observed up to the current iteration, respectively. To comply with the purpose of SO, HASSO considers the primary objective improvement of the associated BBO problems as success, and failure is defined as observing no improvement over the current observed objective value. Initially, the prior distribution is set to be a uniform distribution, with $\alpha = \beta = 1$, which means that HASSO has no prior preference over which arm (hyperparameter) to select. Subsequently, the associated beta distribution of each arm will be updated iteratively to influence the probability of hyperparameter selection. In each iteration, $\alpha=\alpha+1$ when the optimization results in objective improvement (success) or $\beta=\beta+1$ if no improvement is achieved (failure).

\new{
HASSO aims to balance between exploring new hyperparameter configurations and exploiting the best-observed configurations that align with near-optimal regions. By updating the beta distributions based on the actual objective improvements, the method favors exploring multiple promising hyperparameter configurations rather than being restricted to a fixed setting.
The accuracy of surrogate models such as GP, can be greatly affected by the choice of hyperparameters. Suboptimal choices can lead to inaccurate models and subpar optimization performance. 
HASSO addresses this challenge by introducing a stochastic element to mitigate the impact of occasional inaccuracies in the surrogate models due to manual tuning or fixed settings. Instead of solely focusing on surrogate model accuracy which is prone to local optima convergence, HASSO evaluates the overall solution quality of the SO algorithm to assess its effectiveness in achieving the global optimization goal.
}
% Suboptimal hyperparameter settings can result in inaccurate models, and relying solely on manual tuning or using a fixed setting may not yield optimal results. However, HASSO addresses this challenge by adopting a stochastic nature to mitigate the impact of occasional inaccuracies in the surrogate models. Instead, it focuses on assessing the overall solution quality of the SO approach to evaluate its effectiveness in achieving the optimization goals. By considering holistic performance rather than solely relying on surrogate model accuracy, HASSO aims to handle the trade-off between accurate modeling and efficient optimization.

\begin{figure*}[!hbt]
\centering
\subfigure[$\alpha_1=8, \beta_1=4$]{\label{fig:beta1}\includegraphics[width=0.27\linewidth]{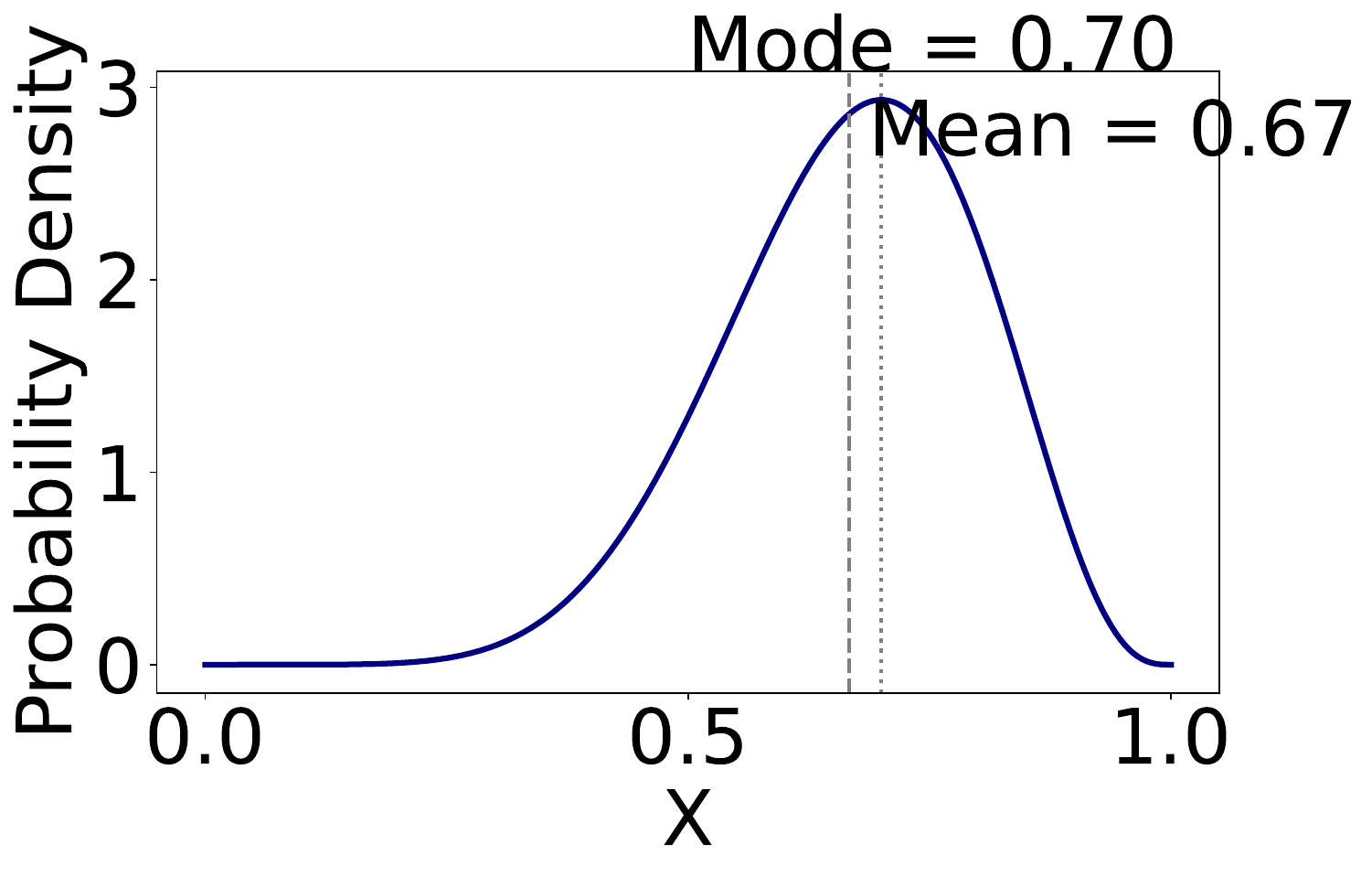}}
\subfigure[$\alpha_2=3, \beta_2=9$]{\label{fig:beta2}\includegraphics[width=0.25\linewidth]{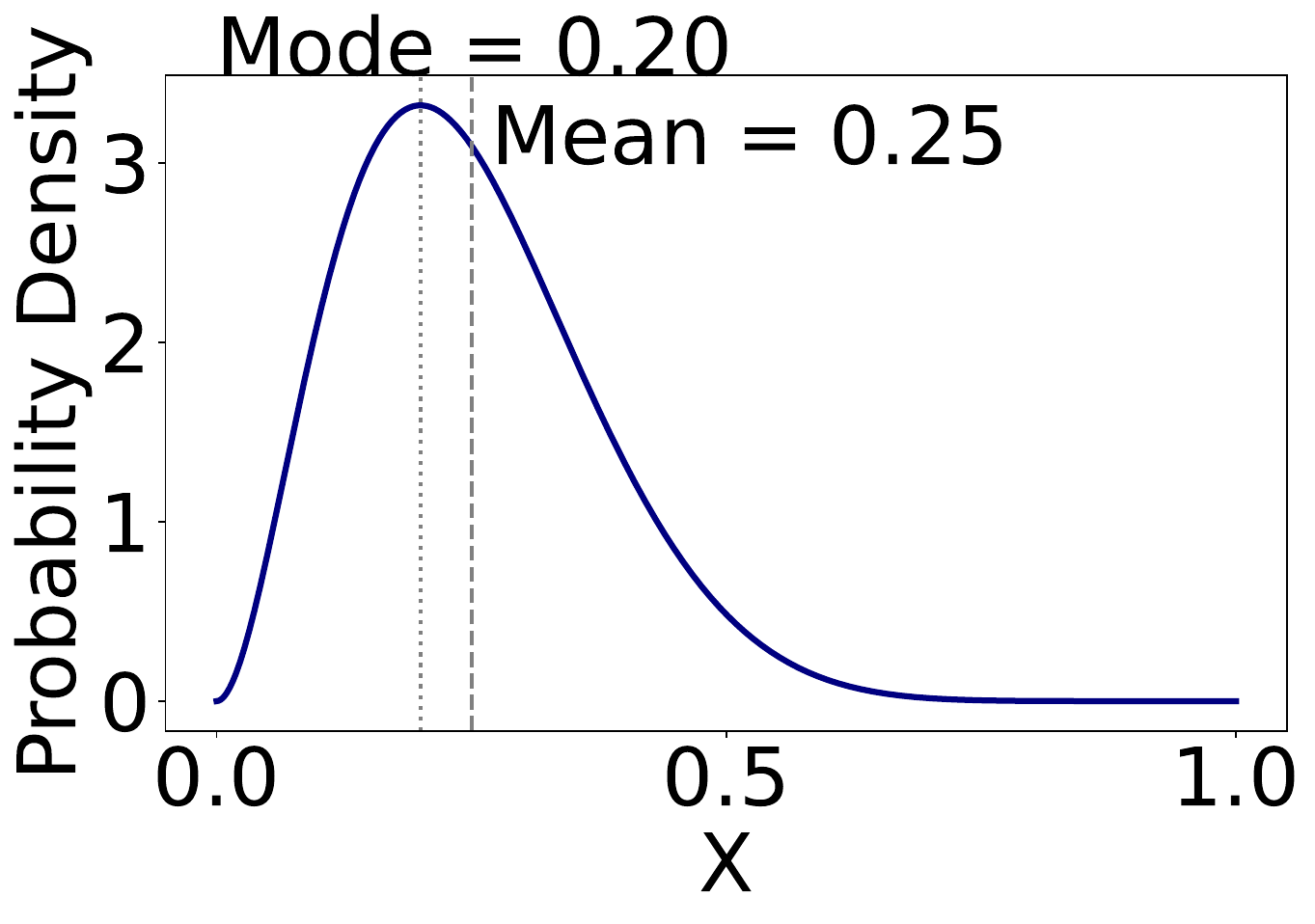}}
\vspace{-1mm}
\caption{Impact of beta distribution on the selection probability of each hyperparameter.} 
\label{fig:beta}
\end{figure*}

Figure~\ref{fig:beta} demonstrates how the posterior beta distributions of two given hyperparameters in HASSO can change after 20 iterations. 
Assuming that the first hyperparameter follows a beta distribution with parameters $(\alpha_1,\beta_1)$, and the second one follows a beta distribution with parameters $(\alpha_2,\beta_2)$. Moreover, each hyperparameter has been selected 10 times for modification ($\alpha_1+\beta_1=\alpha_2+\beta_2$).  Figure~\ref{fig:beta1} shows the probability distribution for the first hyperparameter which resulted in 7 successes (objective improvement) and 3 failures (no improvement). Note that we initialize $\alpha_1=\beta_1=1$; therefore, since we update either value by one per iteration, $\alpha_1=8$ after 7 successes and $\beta_1=4$ after 3 failures. Similarly, Figure~\ref{fig:beta2} shows the probability density for the second hyperparameter which resulted in 2 successes and 8 failures. As the fraction of $\frac{\alpha}{\alpha+\beta}$ increases (more success than failure), the probability distribution is more concentrated towards the upper end of the range and results in a higher mean. On the other hand, if the fraction of $\frac{\beta}{\alpha+\beta}$ increases, the distribution is more spread out and concentrated towards the lower end of the range which results in a lower mean. Comparing the two distributions, HASSO is more likely to select the first hyperparameter due to its considerably higher mean value.

% \vspace{-3mm}

Algorithm~\ref{alg:HASSO} outlines the key steps of our proposed HASSO algorithm, which can be used with any SO approach, given that the acquisition function (e.g., $EI$) and the surrogate model, along with their associated hyperparameters, are provided. The set of algorithmic hyperparameters and their associated ranges should be predefined in an input set $\mathcal{H}$. Each hyperparameter $h_j$ is initialized with a value $\nu_j$, where $a_j\leq \nu_j \leq b_j, \forall j\in \{1,\dots,J\}$.
For each hyperparameter $h_j$, HASSO assigns a beta distribution $B_j$ with initialized shape parameters of $\alpha_j = \beta_j = 1$.
The set of algorithmic hyperparameters used in each iteration $\mathcal{H}^{t}$, is first initialized using $\mathcal{H}$ which holds the best set of values obtained so far for each hyperparameter. Then, the algorithm begins by drawing a random value from the associated beta distribution for each hyperparameter to determine its probability of selection. The arm with the maximum value is then selected for an update, denoted by $k=\argmax_{j \in {1,\dots,J}}u_j$ \new{where $u_j$ is a random sample drawn from the beta distribution for each hyperparameter $j \in \{1,\dots,J\}$}. HASSO can use any updating rule to modify the value of the selected hyperparameter $h_k$ to $\nu_k^{'}$, and stores the set of updated hyperparameters in $\mathcal{H}^t$. 
% \hadis{do we need to say options for updating in this work?} \nazanin{I think that can be clarifed in the experiment section}

% \hadis{talk about Srg-opt SO for alg to optimizing x}\nazanin{you mean adding description for Alg3?}

% The updated hyperparameter value is then used in one of the main parts of the SO algorithm, such as modeling, discretization, or the acquisition function, as it optimizes for $\mathbf{x}$. After selecting the arm with the maximum value, the updated hyperparameter value is used in associated parts of the SO algorithm. 
\new{
After selecting the arm with the maximum value, the updated hyperparameter value is utilized in various components of the SO algorithm.
The updated hyperparameter is passed to step 14 of Algorithm~\ref{alg:HASSO} and subsequently to SO Algorithm~\ref{alg:srgopt} in each iteration through $\mathcal{H}^t$. For instance, if hyperparameter $h_k$ is associated with the surrogate model, its value is updated, and the modified model $M$ is utilized to determine the newly selected candidate point $\mathbf{x}^t$ in step 4 of Algorithm~\ref{alg:srgopt}. 
Moreover, to facilitate the selection of promising candidate points, the solution space is discretized using the hyperparameter values in $\mathcal{H}^t$. 
%The set of unevaluated points in the solution space is denoted as $\mathcal{C}$. 
This discretization is denoted as $\mathcal{C}$ and it represents a set of unevaluated points in the solution space.
The predefined SO strategy, incorporating the updated model $M$, the discretized set of points $\mathcal{C}$, and the predefined acquisition function $AF$, then selects the most promising candidate point $\mathbf{x}^t$ for evaluation in iteration $t$.
In HASSO, improvement ($IMP$) is defined as the difference between the best observed objective value obtained after evaluating the selected candidate $\mathbf{x}^{*(t)}$ and the best objective value obtained in the previous iteration $\mathbf{x}^{*(t-1)}$. 
If the updated hyperparameter setting $\mathcal{H}^t$ leads to the evaluation of a more promising candidate point ($IMP>0$), the algorithm considers it a success. Otherwise, it is considered a failure. 
In the case of successful sampling, HASSO increases the value of $\alpha_k$ associated with the beta distribution of hyperparameter $k$
%. This increase in $\alpha_k$ 
which raises the probability of selecting hyperparameter $k$ in future iterations. Additionally, the updated hyperparameter setting $\mathcal{H}^t$ is stored in the $\mathcal{H}$ set, allowing it to be used in subsequent iterations unless it is modified again.
On the other hand, when there is no objective improvement, HASSO increases the value of $\beta_k$ associated with hyperparameter $k$. By increasing $\beta_k$, the probability of selecting hyperparameter $k$ decreases in future iterations. This adjustment helps HASSO focus on exploring hyperparameter values that may yield better results until a termination condition is met. 
}

\begin{algorithm}[!h]
\caption{{\bf Hyperparameter Adaptive Search for Surrogate Optimization}}
\label{alg:HASSO}
\begin{algorithmic}[1]
\STATE $\mathcal{D}=\{\mathbf{x}^1, \ldots, \mathbf{x}^{n}\}$,  
$\mathcal{F}= \{f(\mathbf{x}^i)|\mathbf{x}^i\in \mathcal{D}\}$, $\mathbf{x}^{*(0)}=\argmin_{\mathbf{x}\in \mathcal{D}} f(\mathbf{x})$, $t=1$, $T$=Total Budget,
\STATE $AF=$ Acquisition function
\STATE $\mathcal{H}=\{h_1=\nu_1,h_2=\nu_2,\dots,h_{J}=\nu_J \vert a_j\leq \nu_j\leq b_j, \forall j \in \{1,\dots,J\} \} $
\STATE $B_{j}\sim Beta(\alpha_{j}=1,\beta_{j}=1) \forall j \in \{1,\dots,J\}$.  
%start the while loop 
\WHILE {$t \leq T$}
%define the best known solution so far 
\STATE $\mathcal{H}^t \leftarrow \mathcal{H}$
\STATE \emph{Choose one arm:} %choose an arm based on beta 
\FOR{$j \in \{1,\dots,J\}$:}
 $u_j$= Draw a random sample from $B_{j}\sim Beta(\alpha_{j},\beta_{j})$ 
\ENDFOR
\STATE $k=\argmax_{j \in \{1,\dots,J\}}u_j$
  %update HP based on arm selection 
\STATE \emph{Update the Hyperparameter:} 
% \STATE $h_{k}^{'}$= Update $h_{k}$ s.t $a_k \leq h_{k} \leq b_k$
% \STATE $\mathbf{\nu}^t=[\nu_1^t,\dots,\nu_J^t]$
\STATE $\nu_{k}^{'}$= Update $h_{k}$ s.t $a_k \leq \nu_{k} \leq b_k$
\STATE $\mathcal{H}^t=\{h_1=\nu_1,h_2=\nu_2,\dots,h_k=\nu_k^{'},\dots,h_{J}=\nu_J\}$
\STATE $\mathbf{x}^t$= Srg-opt($\mathcal{D},\mathcal{F},\mathcal{H}^t, AF$)
 \STATE \emph{Evaluation}: $\mathcal{F}_{\mathbf{x}^t}= f(\mathbf{x}^t)$
\STATE $\mathcal{D}= \mathcal{D}\cup \mathbf{x}^t$; $\mathcal{F}=\mathcal{F}\cup \mathcal{F}_{\mathbf{x}^t}$
\STATE $\mathbf{x}^{*(t)}=\argmin_{\mathbf{x}\in \mathcal{D}} f(\mathbf{x})$ %end of evaluation
%calculate improvmenet 
\STATE $IMP=\mathbf{x}^{*(t-1)}-\mathbf{x}^{*(t)}$
%update HP and alpha, beta based on imp
\IF{$IMP>0$}: 
$\mathcal{H} \leftarrow  \mathcal{H}^t$, $\alpha_{k}=\alpha_{k}+1$
\ELSE: $\beta_{k}=\beta_{k}+1$ 
\ENDIF %end update 
\STATE $\mathbf{x}^{*(t)}=\argmin_{\mathbf{x}\in \mathcal{D}} f(\mathbf{x})$
\STATE $t=t+1$ %iteration 
\ENDWHILE %end while 
\STATE Return $\mathbf{x}^*=\mathbf{x}^{*(T)}$
\end{algorithmic}
\end{algorithm}

\begin{algorithm}[!h] %update algorithm 
\caption{{\bf Srg-opt}}
\label{alg:srgopt}
\begin{algorithmic}[1]
\STATE Input: $\mathcal{D},\mathcal{F},\mathcal{H}, AF$
\STATE $M=$ Fit a surrogate model $\hat{f}$ on $(\mathcal{D},\mathcal{F})$ using related $\mathcal{H}$ hyperparameter values 
\STATE {$\mathcal{C}$= Discretize the solution space using the hyperparameters in $\mathcal{H}$}
% Discretize the solution space using related $\mathcal{H}$ hyperparameter values}
\STATE Determine new candidate point, $\mathbf{x}$, using $M$,$\mathcal{C}$, and $AF$ with hyperparameters in $\mathcal{H}$
\STATE Return $\mathbf{x}$
\end{algorithmic}
\end{algorithm}

% What diffrentiate HASSO from the existing SO approaches is the 
% \nazanin{general discussion for more than 2: exploration-exploitation of the acquisition in our context, number of the times (frequency) that arms are selected}
% \nazanin{after the results, we explain for example for the 2 hyperparameters}

\vspace{-2mm}
\subsection{The Impact of Adaptive Hyperparameter Search on Optimization Performance}

\begin{figure*}[!htb]
\centering
\subfigure{\includegraphics[width=0.45\textwidth]{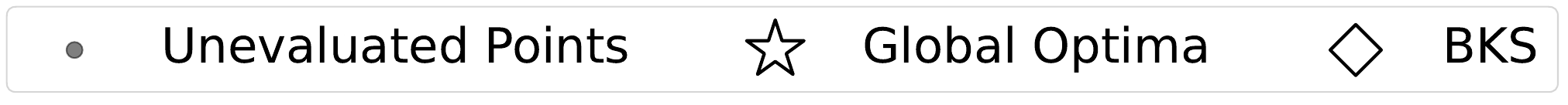}}
\setcounter{subfigure}{0}

\subfigure[Shubert Surface]{\includegraphics[width=0.24\textwidth]{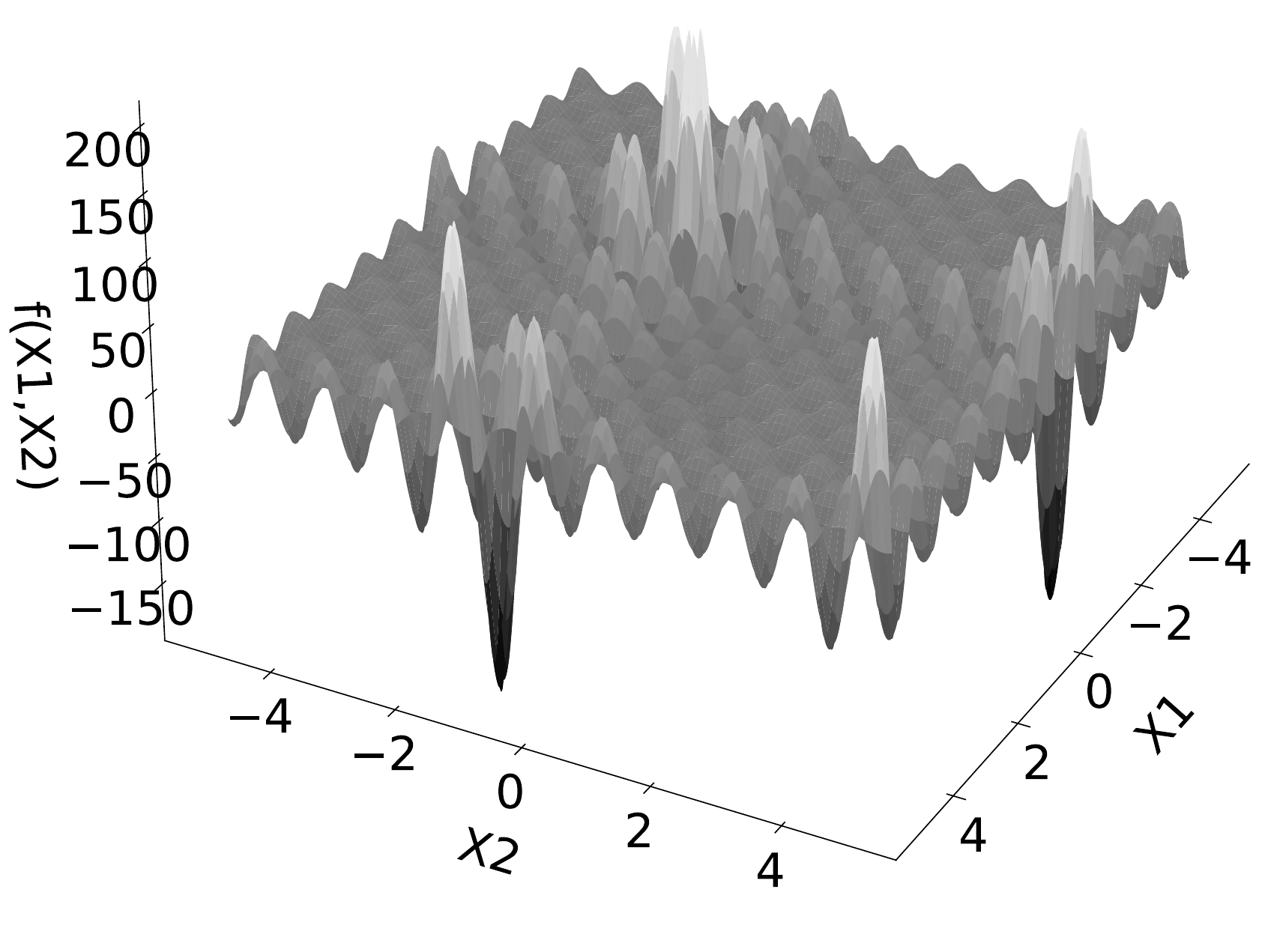}}
\subfigure[SO (10th iteration)]{\includegraphics[width=0.24\textwidth]{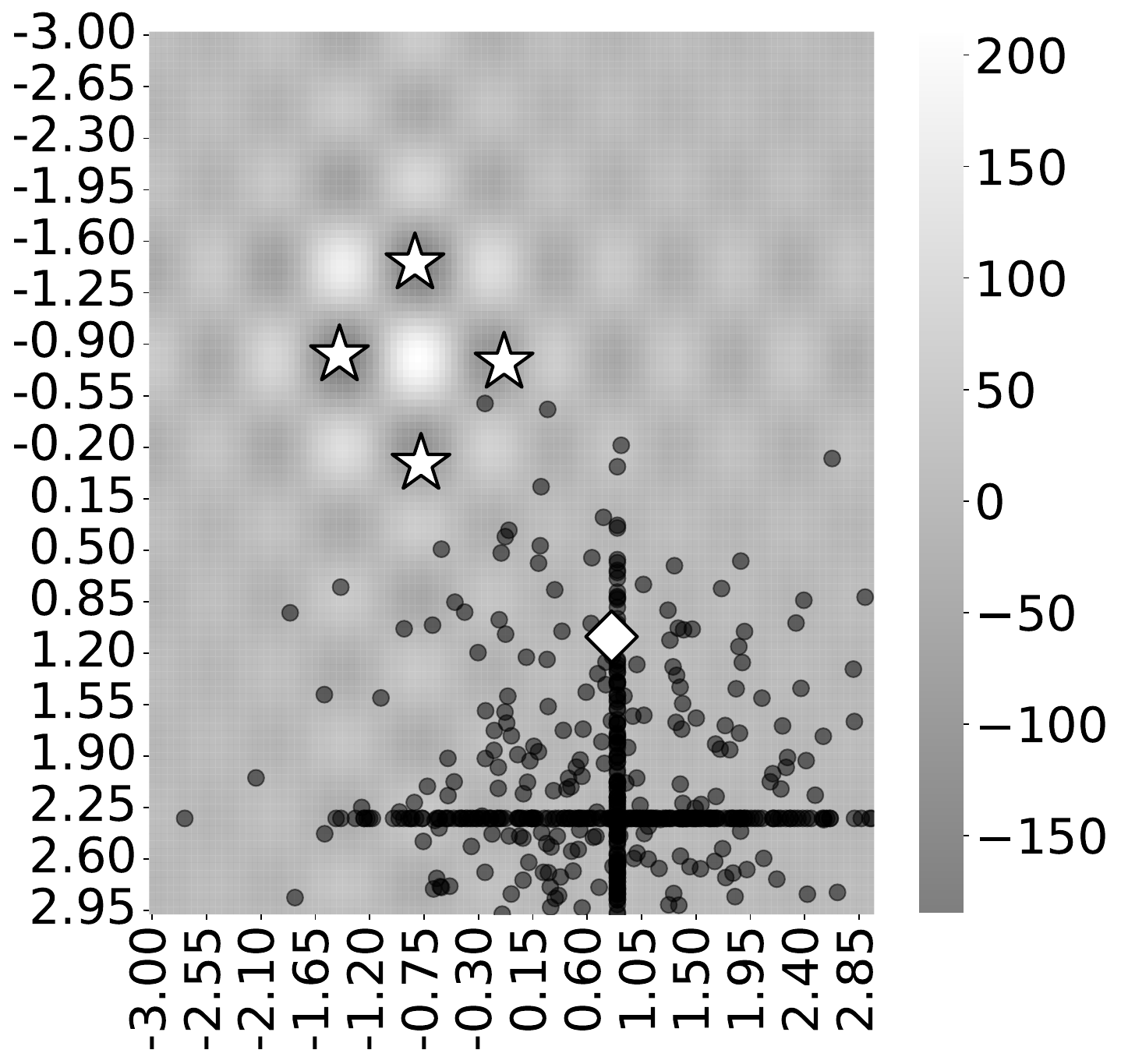}}
\subfigure[SO (50th iteration)]
{\includegraphics[width=0.24\textwidth]{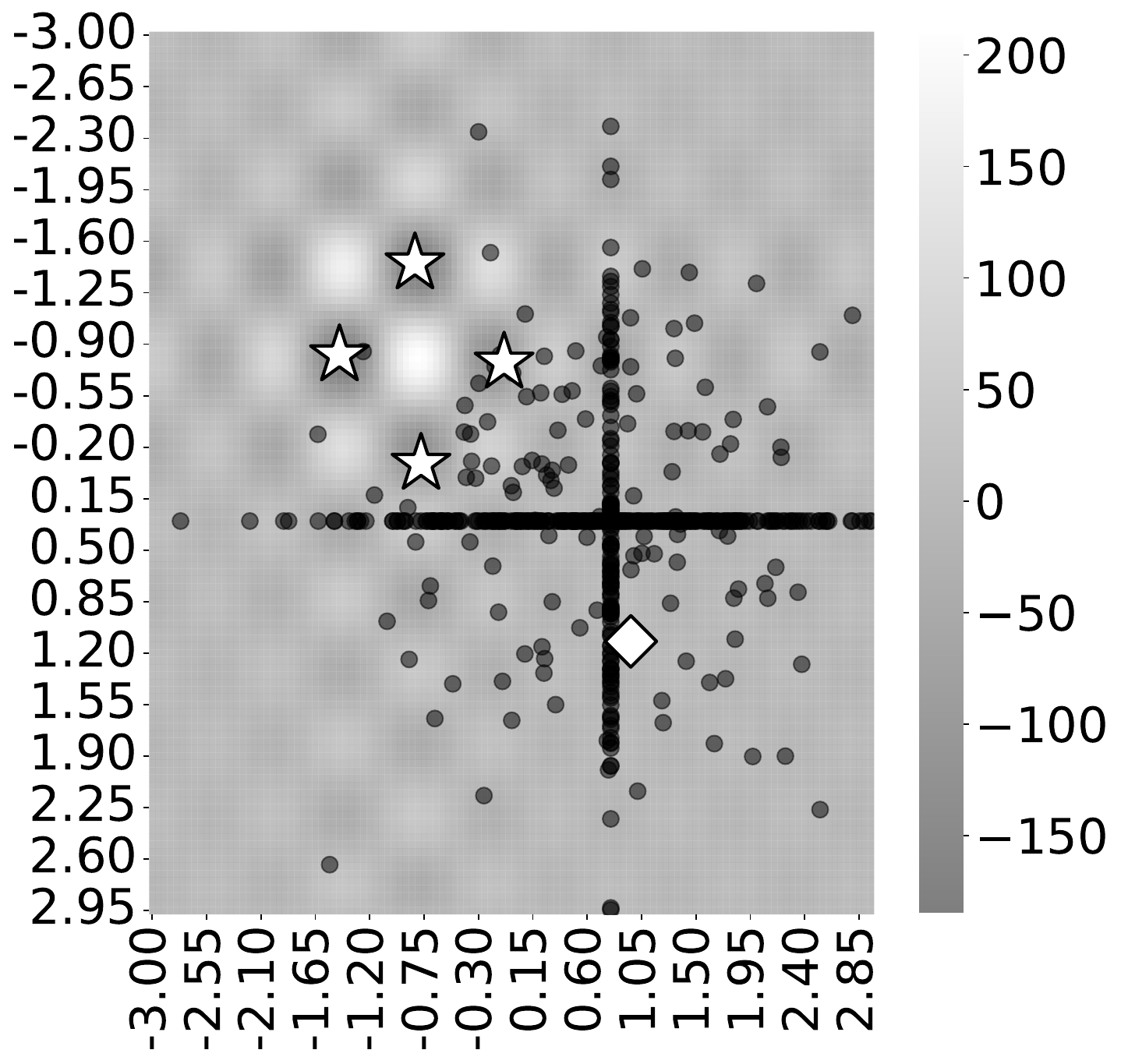}}
\subfigure[SO (100th iteration)]
{\includegraphics[width=0.24\textwidth]{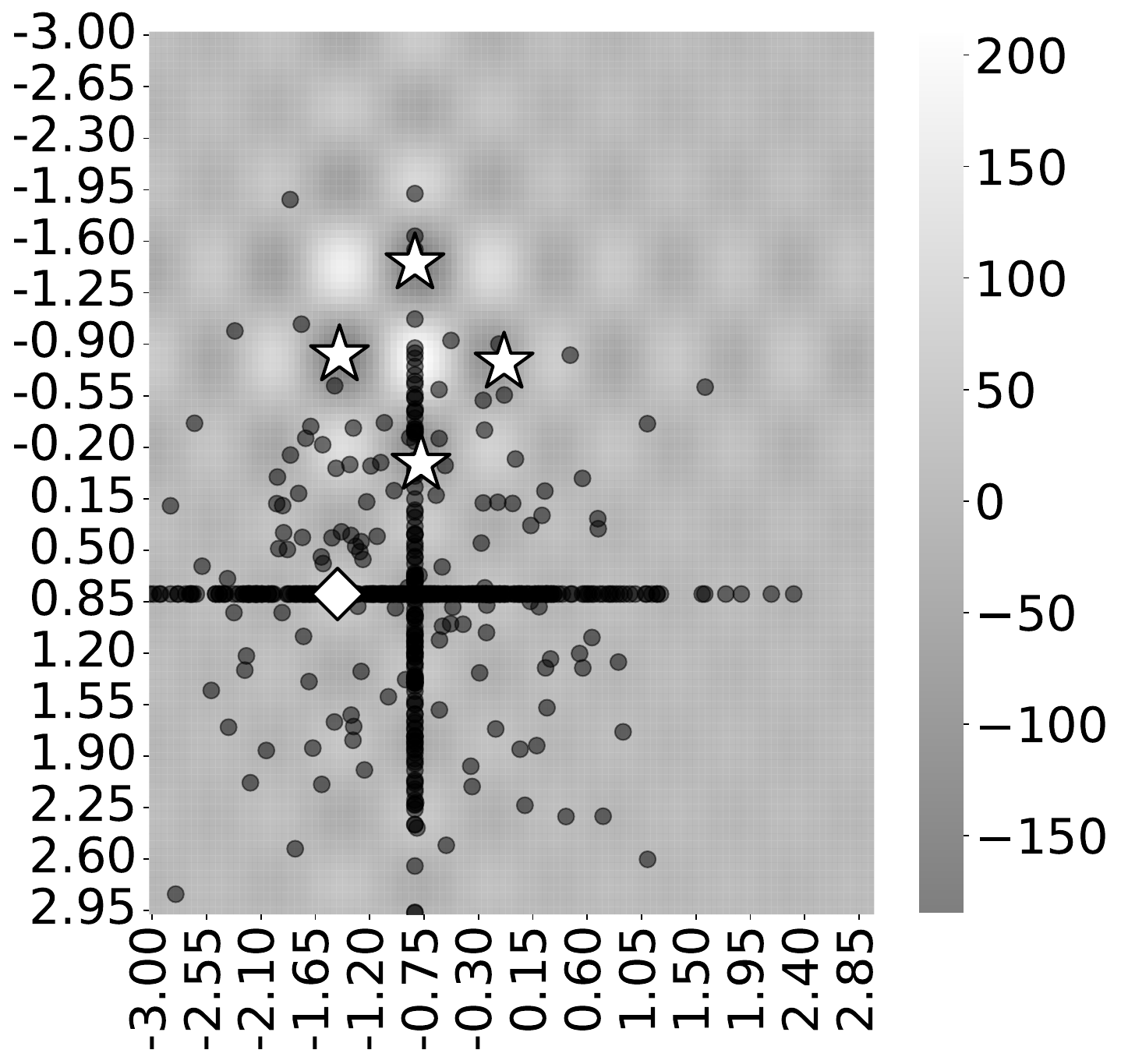}}

\subfigure[Initial State]{\includegraphics[width=0.24\textwidth]{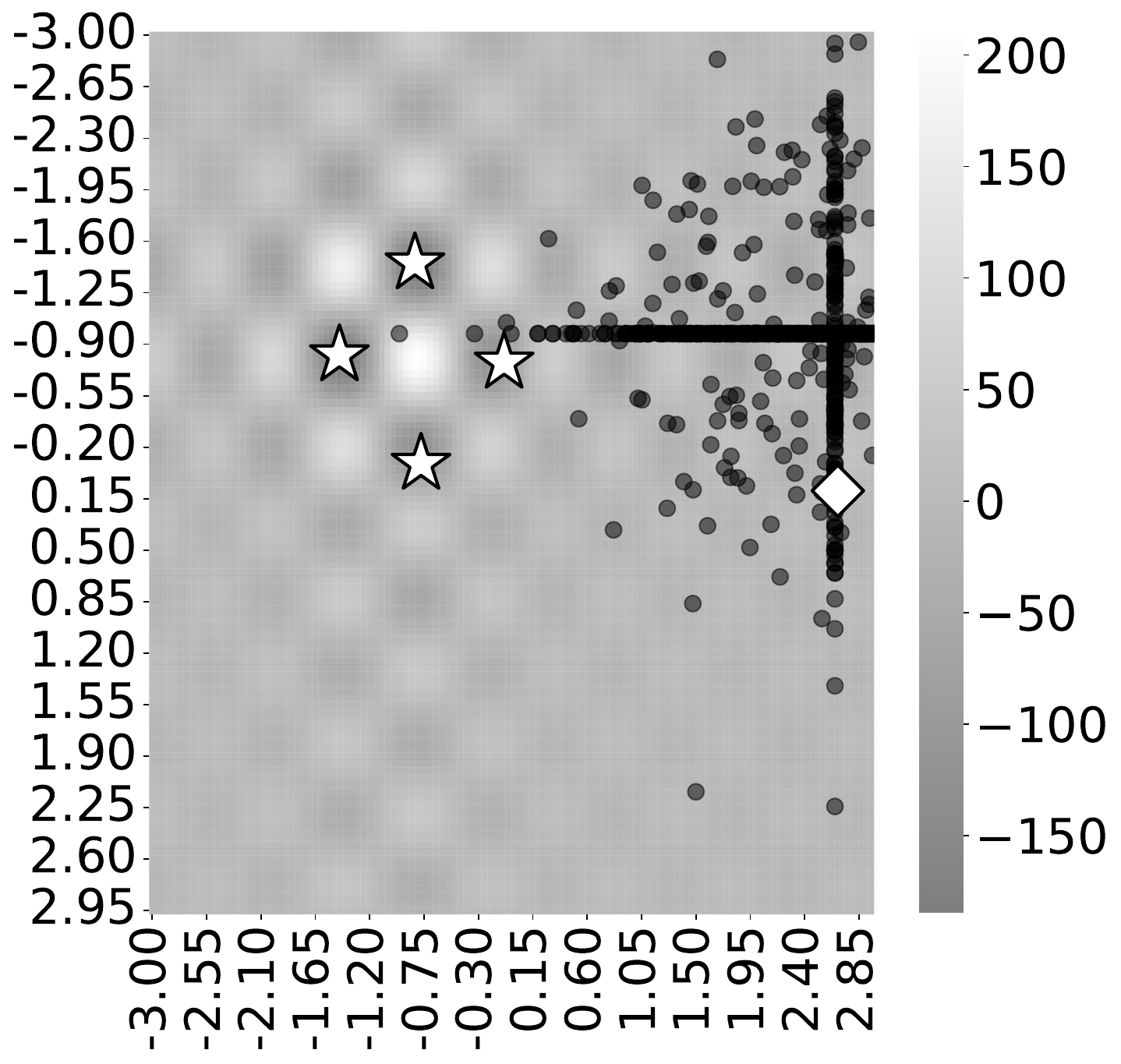}}
\subfigure[HASSO (10th iteration)]{\includegraphics[width=0.24\textwidth]{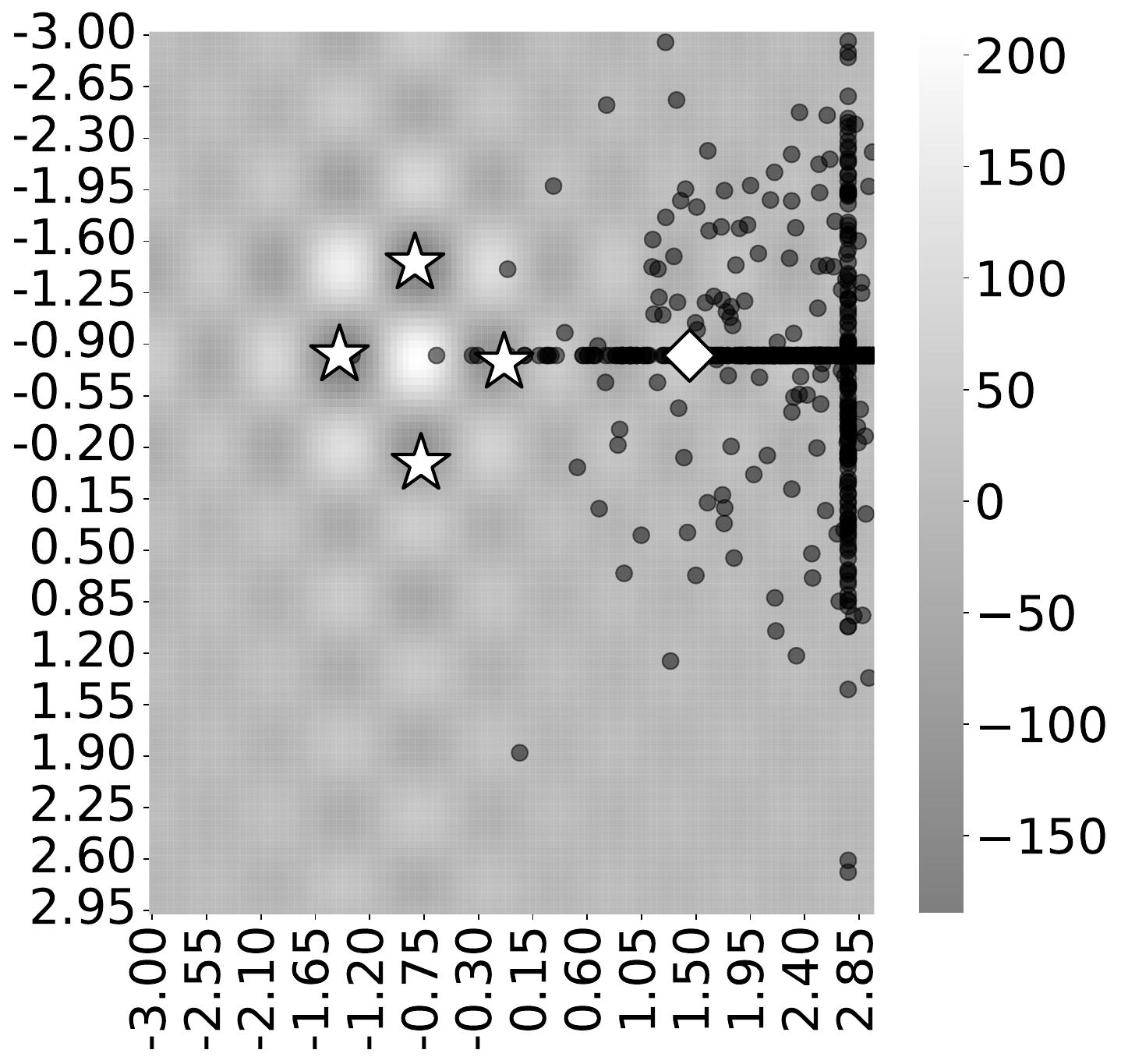}}
\subfigure[HASSO (50th iteration)]
{\includegraphics[width=0.24\textwidth]{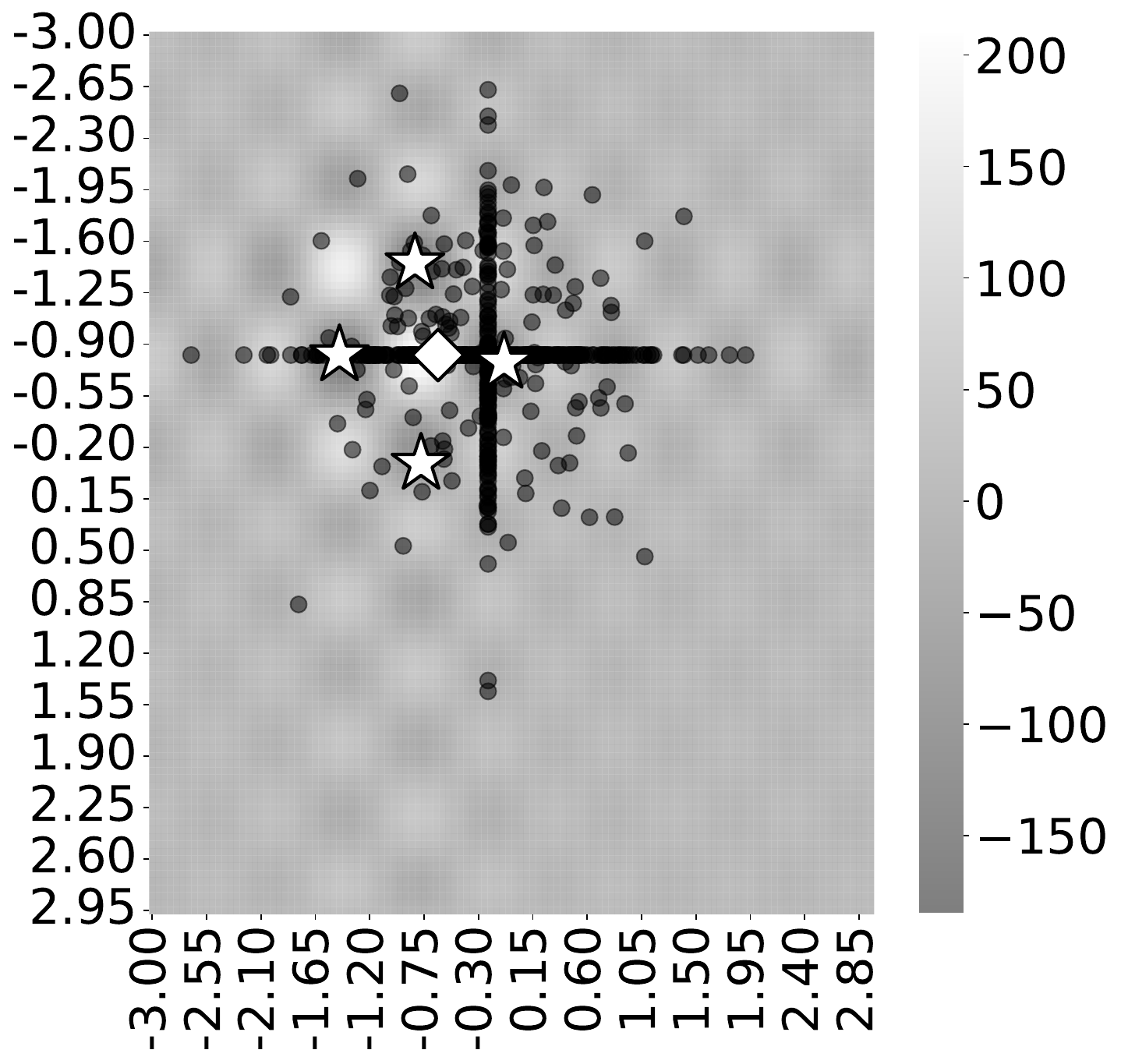}}
\subfigure[HASSO (100th iteration)]
{\includegraphics[width=0.24\textwidth]{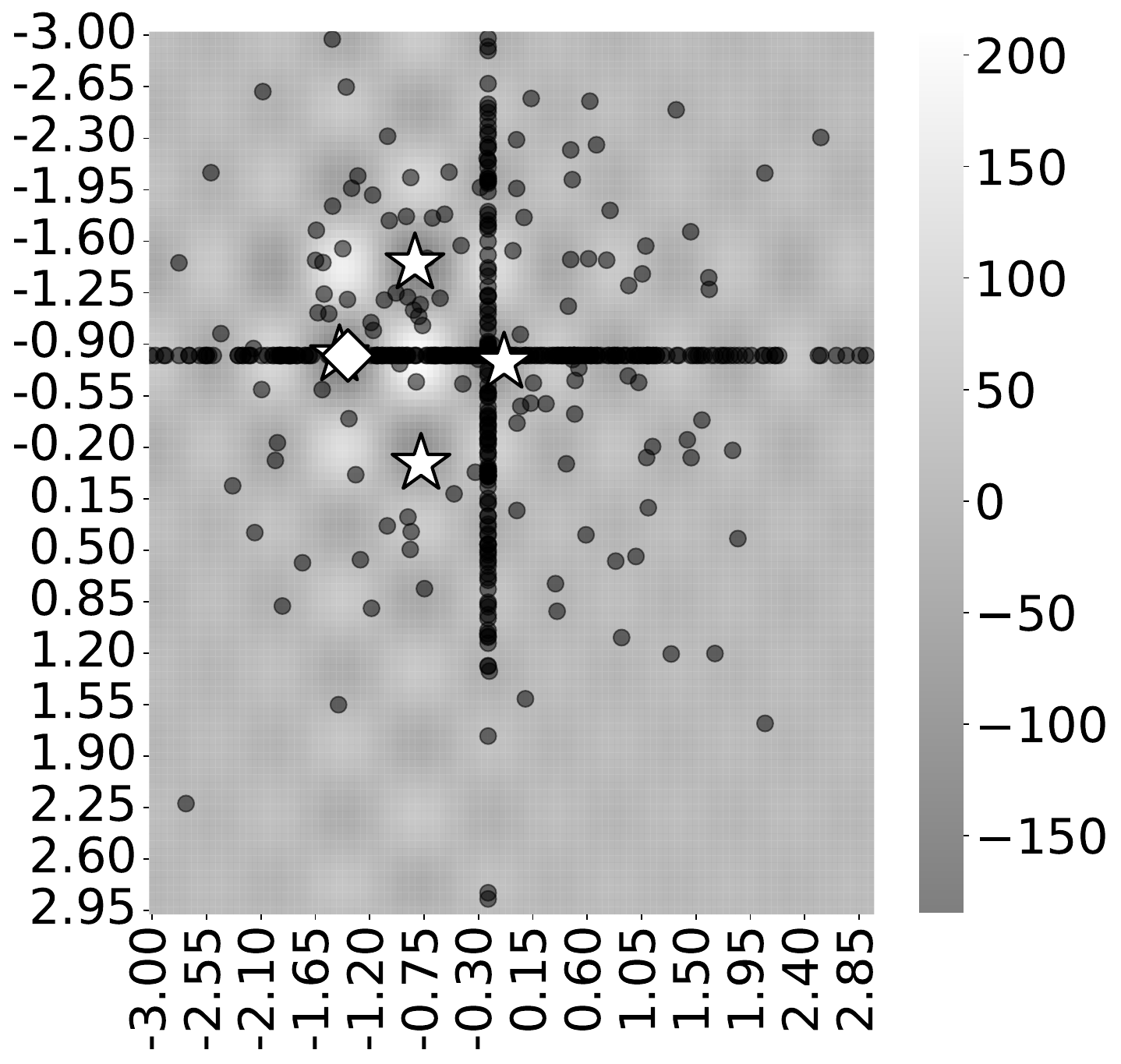}}
\caption{Impact of adaptive hyperparameter Shubert (2$d$).}\label{fig:exp_2D}
%\vspace{-5 mm}
\end{figure*}

In this section, we provide an illustrative example of our proposed adaptive hyperparameter search approach, HASSO, and its effectiveness using the 2-dimensional Shubert global optimization test problem, as depicted in Figure~\ref{fig:exp_2D}. 
Shubert has several local and global minima making it prone to local convergence within a limited budget. 
In Figure~\ref{fig:exp_2D}, HASSO adaptively updates the  discretization radius and the lengthscale of the GP model based on the importance of each hyperparameter in optimizing the Shubert function. 
In contrast, the common practice algorithmic setting, referred to as SO, employs a fixed algorithmic setting for both hyperparameters, primarily focusing on optimizing the Shubert function without employing algorithmic hyperparameter adjustments. The comparison between HASSO and SO reveals that HASSO converges to the near-optimal region faster due to dynamically adjusting the hyperparameter values. Furthermore, HASSO takes into consideration the objective improvement for the Shubert problem when updating the values of hyperparameters in each iteration. It actively explores different hyperparameters based on their contribution to the observed objective value improvement to balance the exploration-exploitation trade-off for hyperparameter tuning. This leads to a strategic movement of the best known solution (BKS) as well as the generated candidate points (Unevaluated Points) toward the desired global optima region. 

% The comparison between HASSO and the common practice that uses a fixed set of algorithmic settings, referred to as SO, clearly reveals that HASSO dynamically adjusts the hyperparameter values to converge to the near-optimal region faster.
% Specifically, in Figure~\ref{fig:exp_2D}, HASSO adaptively updates the radius for discretization and the lengthscale of the GP model based on the importance of each hyperparameter in optimizing the Shubert function. 
% In contrast, SO employs a fixed algorithmic setting for both hyperparameters, primarily focusing on optimizing the Shubert function without employing algorithmic hyperparameter adjustments.

% Furthermore, HASSO takes into consideration the objective improvement for the Shubert problem when updating the values of hyperparameters in each iteration.  For instance, if a certain GP lengthscale value leads to objective improvement, HASSO utilizes that value until it discovers another lengthscale value that improves the observed objective value further. Therefore, HASSO actively explores different hyperparameters based on their contribution to the observed objective value improvement to balance the exploration-exploitation trade-off for hyperparameter tuning. 

% Thus, HASSO not only balances the exploration-exploitation trade-off in the acquisition criterion, similar to other standard SO approaches but also in the algorithmic setting by actively exploring each hyperparameter value based on its contribution to the main objective.

\section{EXPERIMENTS}\label{sec:result}

% In order to expound upon the efficacy of our proposed adaptive search method, \emph{HASSO}, we have undertaken empirical analysis utilizing established surrogate optimization algorithms on a diverse range of global optimization test problems, encompassing both low and high-dimensional settings.
In this section, we present the results of our empirical analysis evaluating the effectiveness of the proposed adaptive search method, \emph{HASSO}, on a diverse range of global optimization test problems. 
%we present the results of our empirical analysis of the efficacy of the proposed adaptive search method, \emph{HASSO}, on a diverse range of global optimization test problems. 
To evaluate the performance of HASSO, we utilize established SO algorithms. The test problems cover both low and high-dimensional settings, and we examine the performance of HASSO across different problem dimensions and search budgets. We also analyze the behavior of HASSO by investigating the impact of different initializations on its performance.

\textbf{Baselines.}
% The following paragraph presents a summary of the baseline scenarios considered in the experiment section.
The experiment section includes several baseline scenarios that were considered in our study. In the first scenario, denoted as \emph{SO}, the surrogate optimization algorithm employs a fixed initialization for all hyperparameters. The second scenario, \emph{SO-2}, is based on the r-rule, as proposed by \citeN{regis2013combining}, to update the radius for candidate generation while all other hyperparameters are fixed. The third scenario, \emph{SO-grid}, involves designing a grid based on the range of each hyperparameter and searching for the best setting under each. 
% This scenario can be referred to as a multi-level hyperparameter search, as the best hyperparameter values are defined hierarchically in each iteration. 
Finally, the fourth scenario, \emph{SO-rand}, performs random updates on all hyperparameters in each iteration.
% The first scenario, \emph{SO}, represents the regular surrogate optimization algorithmic structure with fixed initialization for all hyperparameters. The second scenario, \emph{SO-2}, is inspired by the work of \citeN{regis2013combining} and utilizes the r-rule to update the radius for candidate generation while all other hyperparameters are based on fixed initialization. The third scenario, \emph{SO-grid}, designs a grid based on the range of each hyperparameter and searches for the best setting under each. This scenario can also be referred to as a multi-level hyperparameter search since the best hyperparameter values are defined hierarchically in each iteration. The fourth scenario, \emph{SO-rand} performs random updates on all hyperparameters in each iteration. 
{We examine two alternative strategies for the HASSO algorithm, achieved by modifying the hyperparameter updating strategy as described in Algorithm~\ref{alg:HASSO}. The first strategy, denoted as \emph{HASSO-decay}, employs a decay function to update the hyperparameters based on the trade-off between exploration and exploitation. Specifically, the \emph{HASSO-decay} strategy adjusts the hyperparameter values differently for positive and non-positive improvement, aiming to promote more exploitation (e.g., by reducing the lengthscale value in the GP kernel) when positive improvement is observed, and more exploration (e.g., by increasing the radius) when no improvement is achieved. As the optimization process proceeds, the decay function reduces the magnitude of the updates to the hyperparameters. In contrast, the second strategy, denoted as \emph{HASSO-rand}, uses only random updates for each hyperparameter, within its predefined range. The unrestricted updates considered by \emph{HASSO-rand} allow the algorithm to explore a wider range of hyperparameter values, potentially enhancing the overall search performance.
% We examine two distinct strategies for the HASSO algorithm, achieved by modifying the updating strategy as described in Algorithm~\ref{alg:HASSO}. The first strategy, \emph{HASSO-decay}, employs a decay function to update the hyperparameters based on the objectives of exploration and exploitation. In particular, the \emph{HASSO-decay} strategy uses positive improvement to facilitate more exploitation (for instance, reducing the lengthscale value in the GP kernel) and no improvement to promote more exploration (for instance, increasing the radius) on the hyperparameter values. As the optimization process progresses, the decay function induces fewer alterations to the hyperparameter values. On the other hand, the second strategy, \emph{HASSO-rand}, only utilizes random updates, which are based on the range of each hyperparameter. The unrestricted updates considered by \emph{HASSO-rand} allow for exploring a broader range of values for each hyperparameter, which can enhance the overall search process.
}
The summary of all scenarios is provided in Table~\ref{tab:baselines}.

% Finally, the fifth scenario, \emph{HASSO-decay} \hadis{this is not a baseline you should say we use two different update strategies for HASSO random and decay and introduce them, revise this par}, is the closest to HASSO, as it adopts the idea of Thompson Sampling and beta distribution to select the hyperparameters. However, it uses a decay function to update the hyperparameters based on exploration and exploitation objectives. For the updates, \emph{HASSO-decay} considers positive improvement to perform more exploitation (e.g., lower lengthscale value in GP kernel) and no improvement to perform more exploration (e.g., larger radius) on the hyperparameter values. The decay function results in fewer changes in hyperparameter values as the optimization process progresses. 

\begin{table}[htbp]
  \centering
  \caption{Experiment baselines comparison.}
  \begin{adjustbox}{width={0.7\textwidth}}
    \begin{tabular}{|l|l|l|l|}
    \hline
    %\toprule
    Algorithm & Srg hyperparameter  & Dis hyperparameter & Update  \\
    \hline
    %\midrule
    SO    & fixed  & fixed  & None  \\
    \hline
    %\midrule
    SO-2  & fixed  & r-rule  & factor 2  \\
    \hline
    %\midrule
    SO-grid  & grid-search to find best  & grid-search to find best  & use the best  \\
    \hline
    %\midrule
    SO-rand & random change & random change  & random  \\
    \hline
    %\midrule
    HASSO-decay & adaptive decay change & adaptive decay change & decay function  \\
    \hline
    %\midrule
    HASSO-rand & adaptive random change & adaptive random change & random  \\
    \hline
    %\bottomrule
    \end{tabular}%
  \label{tab:baselines}%
  \end{adjustbox}
\end{table}%

\textbf{Experiment Set-up:}
In this study, we investigate the performance of the considered baselines in solving three global optimization test problems that possess diverse shapes and topological characteristics in both medium and high-dimensional settings. To evaluate each test problem, we employ the Latin Hypercube Design method with the maximin criterion to generate distinct sets of initial points, each consisting of $2*(d+1)$ evaluations. We repeat the experiments 30 times, using each set of initial points. In medium-dimensional experiments, the budget is set to 400 and the discretized solution space size is 1000. In medium-dimensional problems, the budget is set to 500 and the discretized solution space size is 4000. 

\subsection{Empirical Results}
\begin{figure}[!h]
\centering
\subfigure{\includegraphics[width=0.8\linewidth]{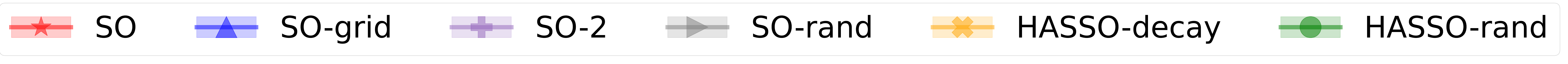}}
\setcounter{subfigure}{0}
\subfigure[UCB-RosenBrock (30$d$)]{\label{exp:Rosenbrock30}\includegraphics[width=0.25\linewidth]{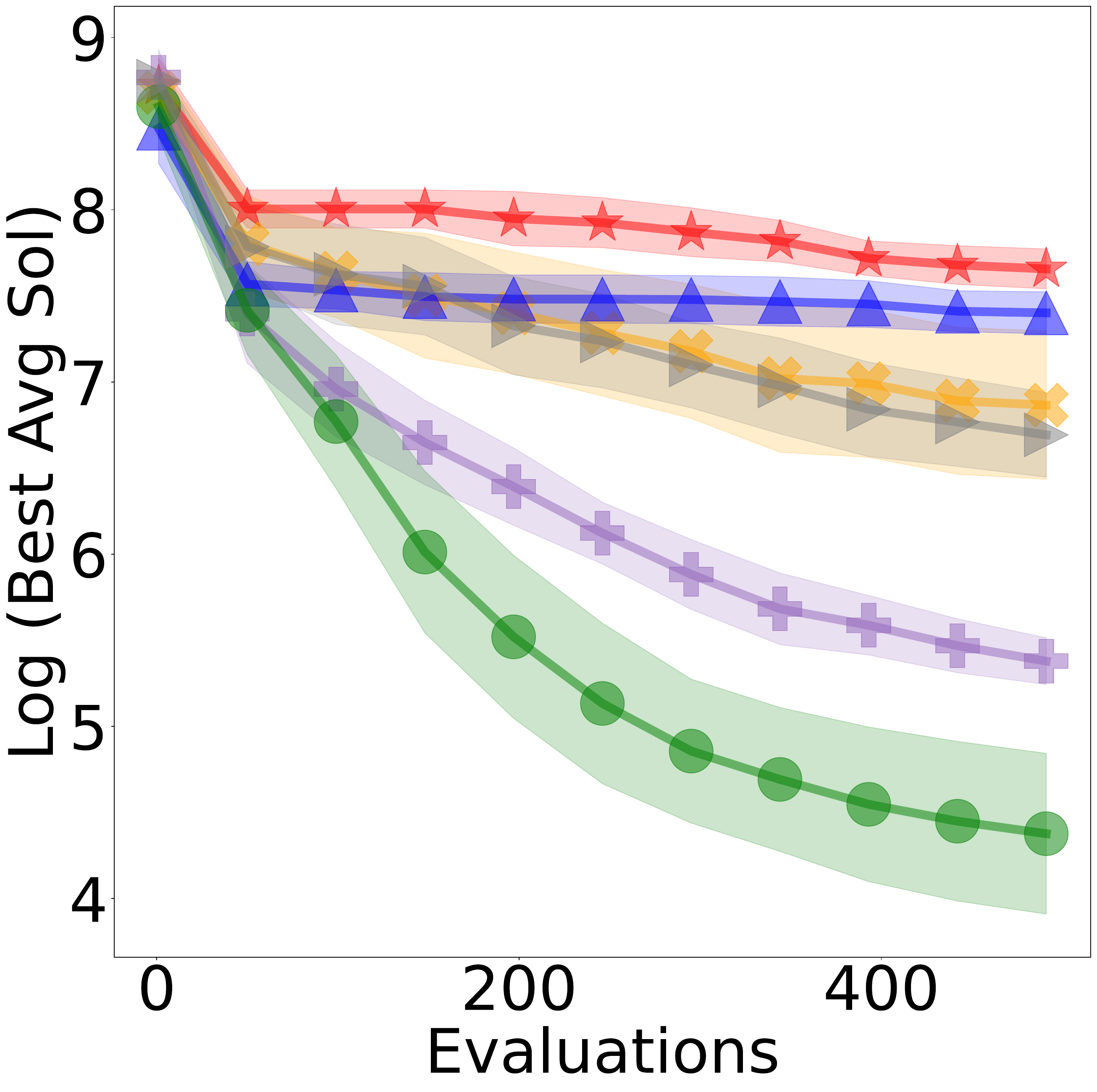}}
\subfigure[Wscore-Rastrigin (30$d$)]{\label{exp:Rastrigin30}\includegraphics[width=0.27\linewidth]{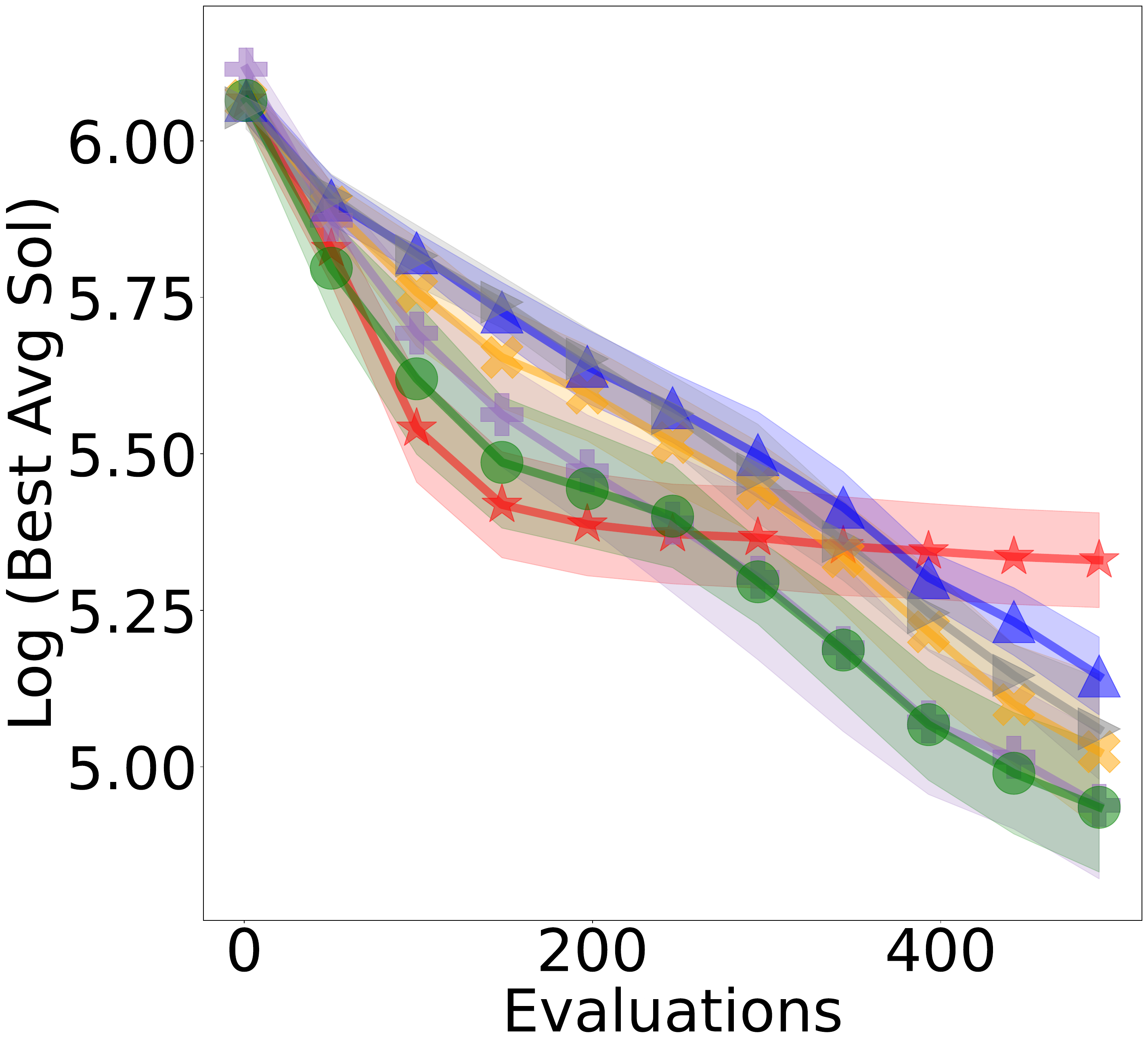}}
\subfigure[EI-Perm (30$d$)]{\label{exp:Perm30}\includegraphics[width=0.25\linewidth]{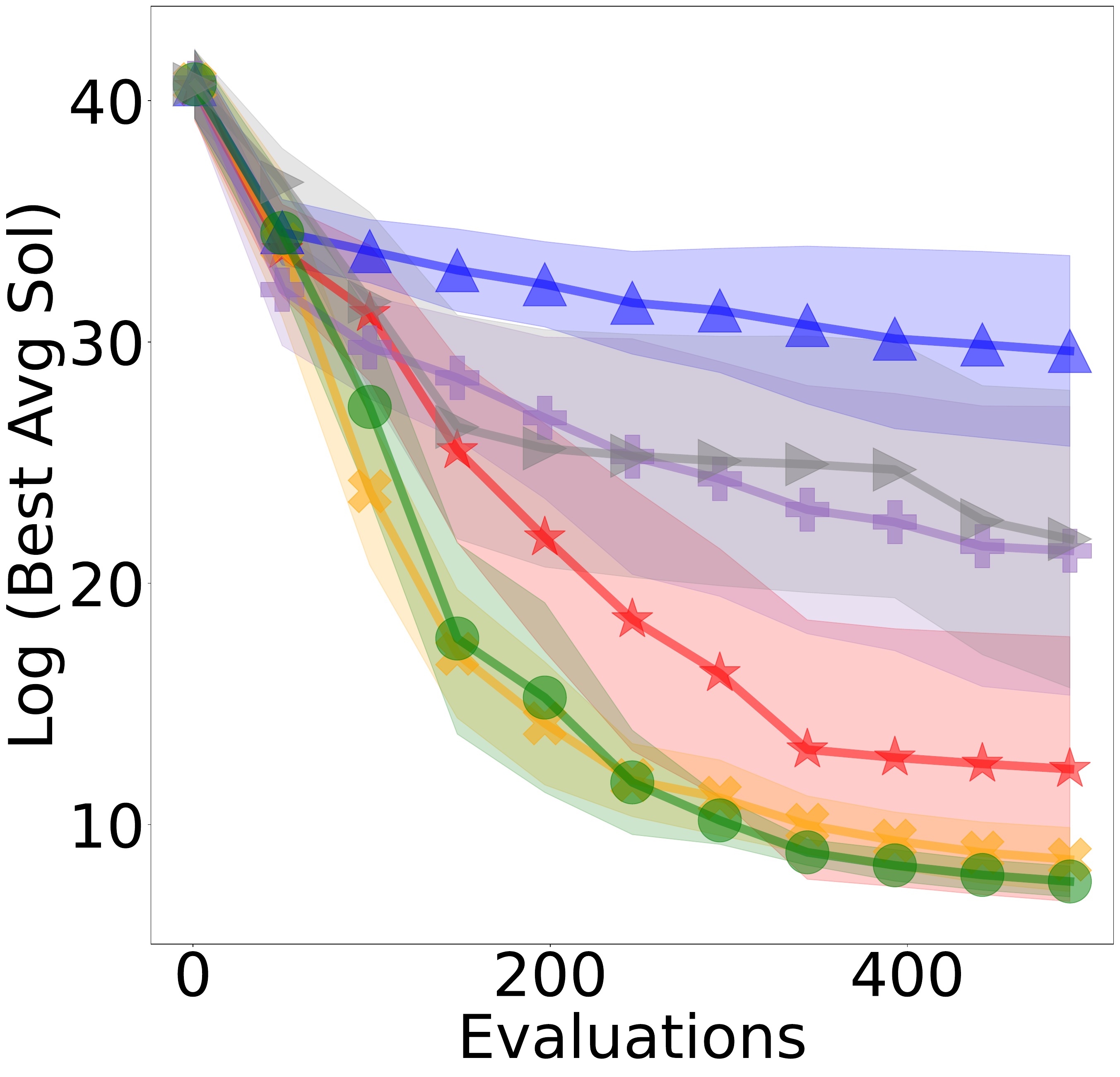}}

\subfigure[UCB-RosenBrock (10$d$)]{\label{exp:Rosenbrock10}\includegraphics[width=0.25\linewidth]{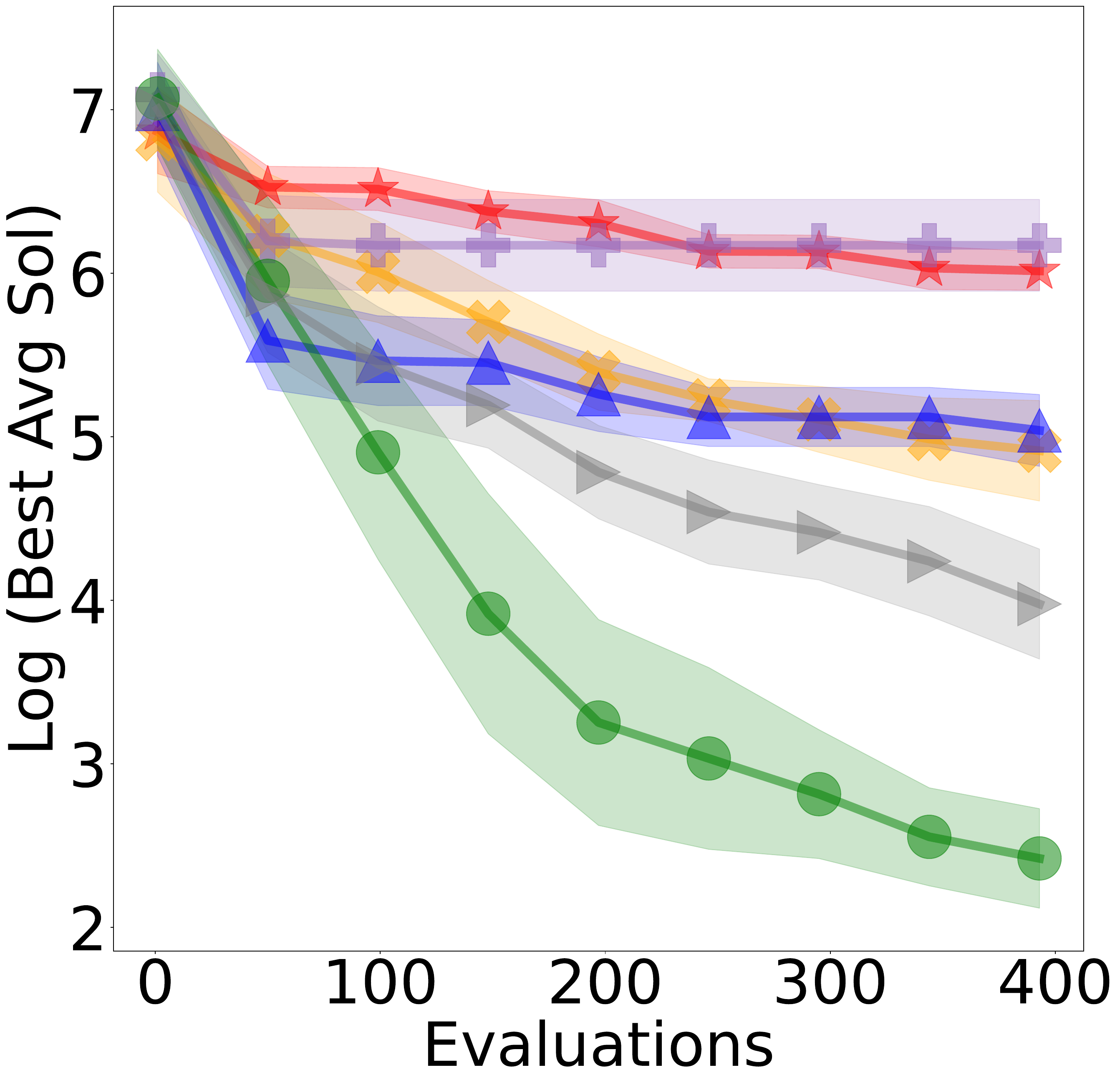}}
\subfigure[Wscore-Rastrigin (10$d$)]{\label{exp:Rastrigin10}\includegraphics[width=0.27\linewidth]{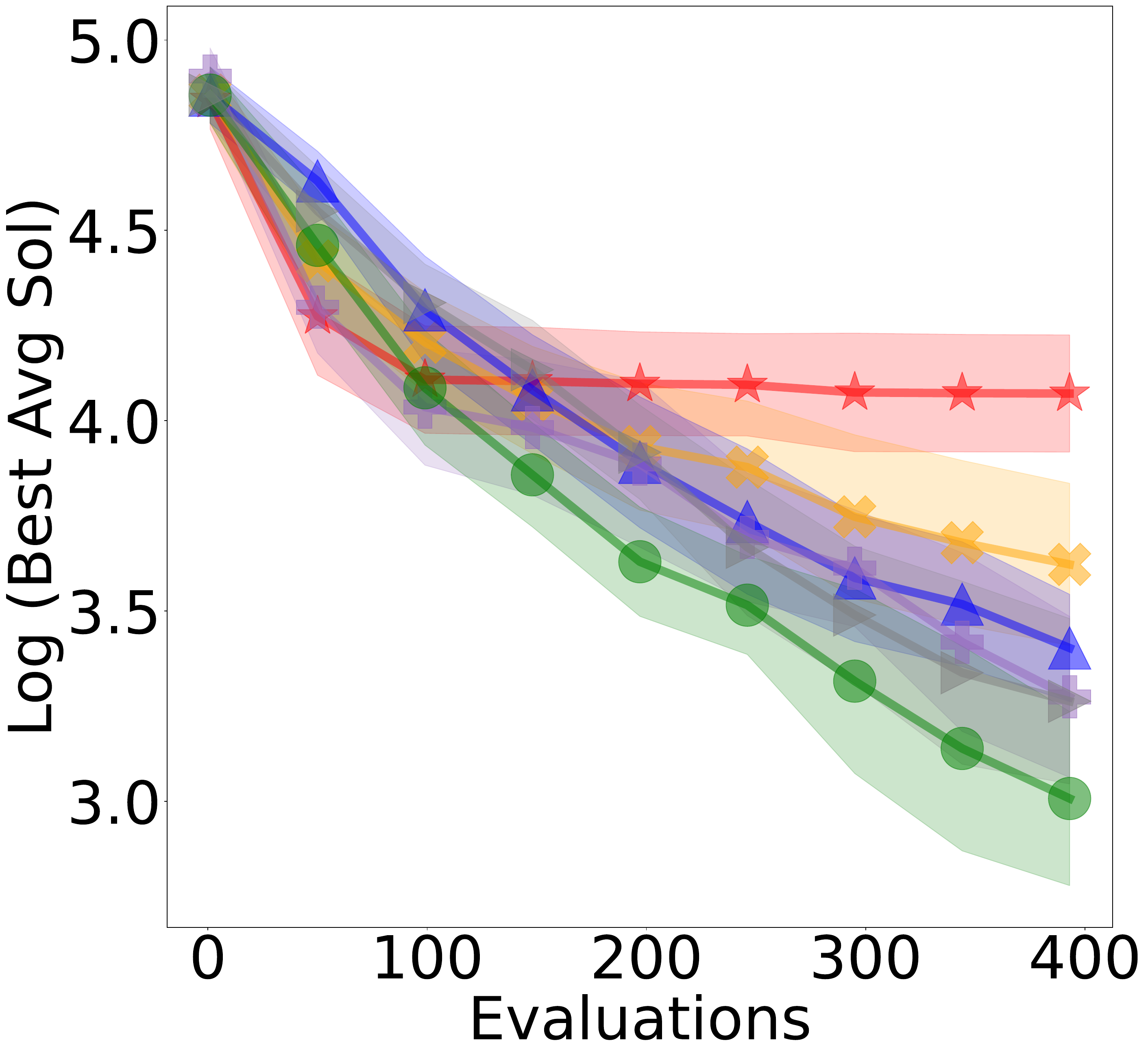}}
\subfigure[EI-Perm (10$d$)]{\label{exp:Perm10}\includegraphics[width=0.25\linewidth]{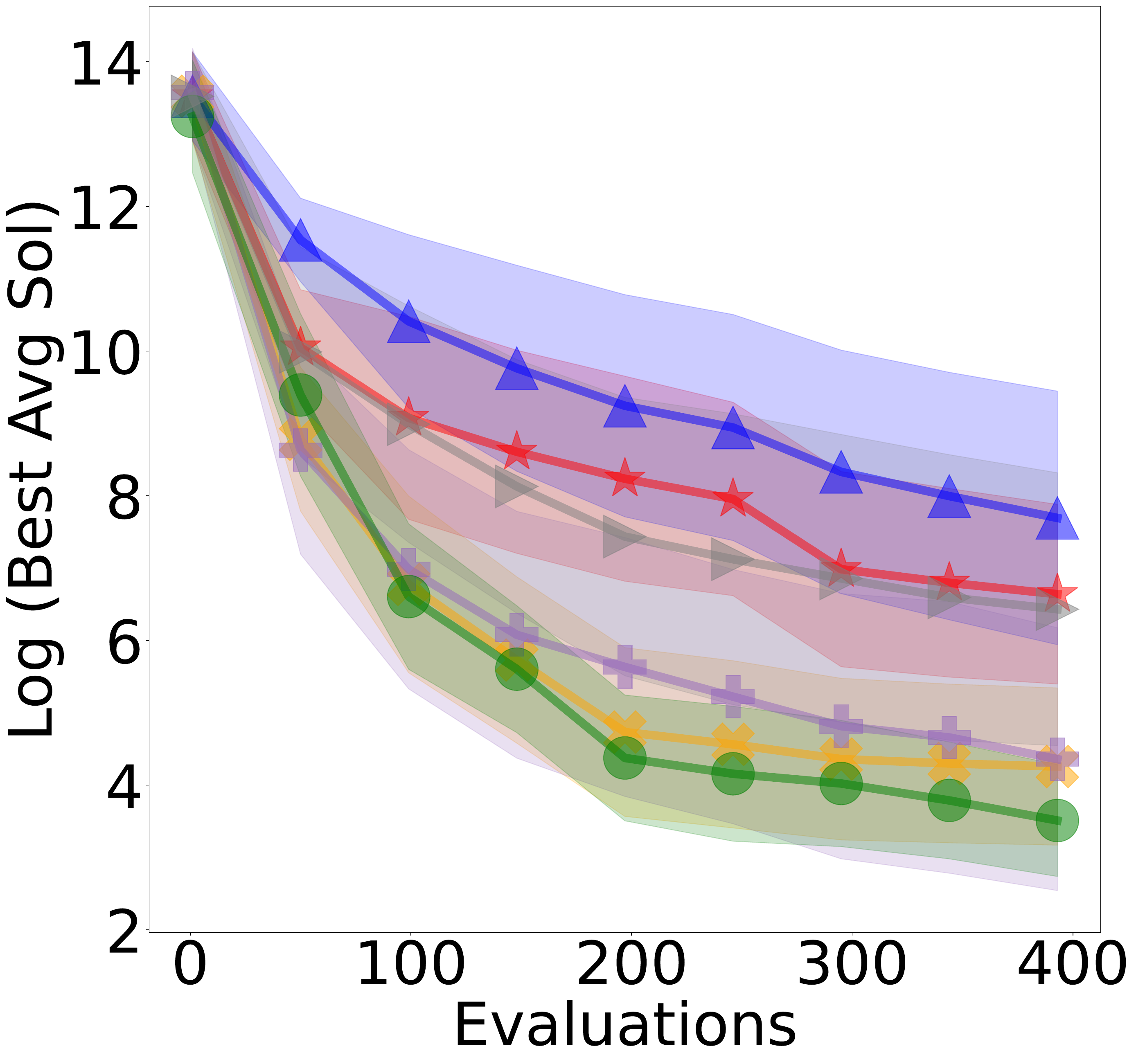}}
\caption{Performance comparison of HASSO and other baselines for different BBO algorithms and global optimization test problems.}
\label{exp:final}
% \vspace{-5 mm}
\end{figure}

Figure~\ref{exp:final} presents the results of our empirical analysis, which compares the performance of our proposed method, HASSO, with other
% Figure~\ref{exp:final} illustrates the experimental results obtained by our proposed approach, HASSO, and its comparison with other 
baselines. We evaluated our method on three global optimization problems that exhibit different topological characteristics across both mid-dimensional and high-dimensional settings.
\new{ 
Rosenbrock is a unimodal valley-shaped function of the form $\sum_{i=1}^{d-1}[100(x_{i+1}-x_i^2)^2+(x_i-1)^2]$, Rastrigin is a multimodal function of the form $10d+\sum_{i=1}^{d}[x_i^2-10cos(2 \pi x_i)]$, and Perm is a bowl-shaped function of the form $\sum_{i=1}^{d} ( \sum_{j=1}^{d}{ (j+\beta)(x_j^{i}-\frac{1}{j^i}))^{2} }$. 
}
% We have conducted our experiments using three global optimization problems that possess distinct topological characteristics in both mid-dimensional and high-dimensional settings. 
To solve these problems, we employed commonly used SO algorithms, namely UCB, Wscore, and EI. 
%\hadis{so you do not have PI in the results? if not we should remove it from the background section} \nazanin{yes, I implemented but did not put the results}.
We also considered two hyperparameters, namely the lengthscale for the Matern kernel in the GP model, and the radius used for dynamic candidate generation.
% To find the global optima of these test problems, we have employed popular SO approaches, including UCB, Wscore, and EI. Additionally, we have considered two hyperparameters, namely, the length scale for Gaussian Process's Matern kernel and the radius for dynamic candidate generation.

% Our analysis in Figure~\ref{exp:final} reveals that using fixed hyperparameter values within SO algorithms, which is a common practice, can put the approach at a disadvantage when compared with using random updates (SO-rand) in most test problems. Therefore, incorporating random updates in each iteration can lead to better results compared with a fixed initialization.

Based on the experimental results presented in Figure~\ref{exp:final}, we observed that using fixed hyperparameter values in SO algorithms can lead to suboptimal results compared to using random updates (SO-rand) for most test problems. Incorporating random updates in each iteration can lead to faster convergence and better results.
In contrast, defining a fixed discrete grid for each hyperparameter and searching for the best setting hierarchically (SO-grid) might be worse than the fixed setting (SO) for smooth bowl-shaped problems with a unique global optimum, such as Perm. 
% Moreover, defining a fixed discrete grid for each hyperparameter and looking for the best setting in a hierarchical manner (SO-grid) might be worse than the fixed setting (SO) for smooth bowl-shaped problems with a unique global optimum, such as Perm.
The SO-2 strategy, which applies the r-rule update in a fixed setting, was competitive with the SO, SO-grid, and SO-rand baselines. 
% Comparing the performance of the SO-2 strategy, which applies the r-rule update in a fixed setting, we observe that it is competitive with SO, SO-grid, and SO-rand baselines. 
Furthermore, we found that the HASSO-decay performance is competitive with HASSO-rand for the smooth Perm test problem. However, the decay in the updates puts HASSO-decay at a significant disadvantage for test problems that require more exploration, such as Rosenbrock and Rastrigin, especially with increasing dimensionality.
% Furthermore, the HASSO-decay performance is competitive with HASSO for the smooth Perm test problem. However, the decay in the updates puts it at a significant disadvantage for test problems that require more exploration, such as Rosenbrock and Rastrigin. This disadvantage further intensifies with increasing dimensionality.
% Overall, our proposed approach, HASSO, outperforms all considered test problems. The results indicate that HASSO can be highly effective in global optimization problems with diverse topological characteristics, offering a promising alternative to the conventional fixed-hyperparameter approach.
\new{We found that HASSO-rand performs better than other baselines in all the test problems we considered.
Specifically, HASSO is highly effective in global optimization problems with diverse topological characteristics.}
% and offers a promising alternative to the fixed-hyperparameter approach. 

\subsection{HASSO: Analysis on the Updating Rule}
\begin{figure}[!h]
\centering
\subfigure[UCB-RosenBrock(10$d$)]{\label{hasso-ucb}\includegraphics[width=0.24\linewidth]{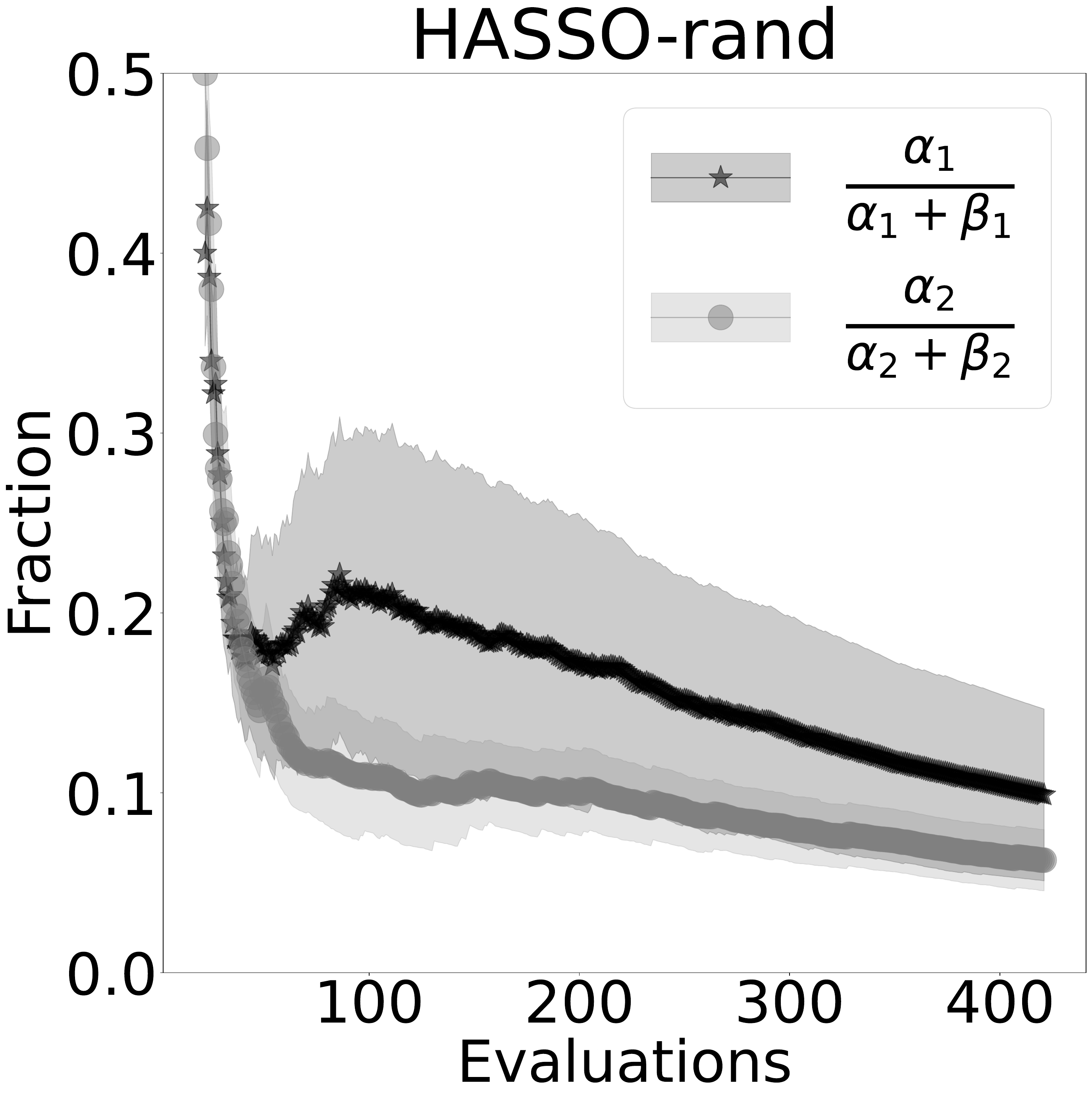}}
\subfigure[UCB-RosenBrock(10$d$)]{\label{decay-ucb}\includegraphics[width=0.24\linewidth]{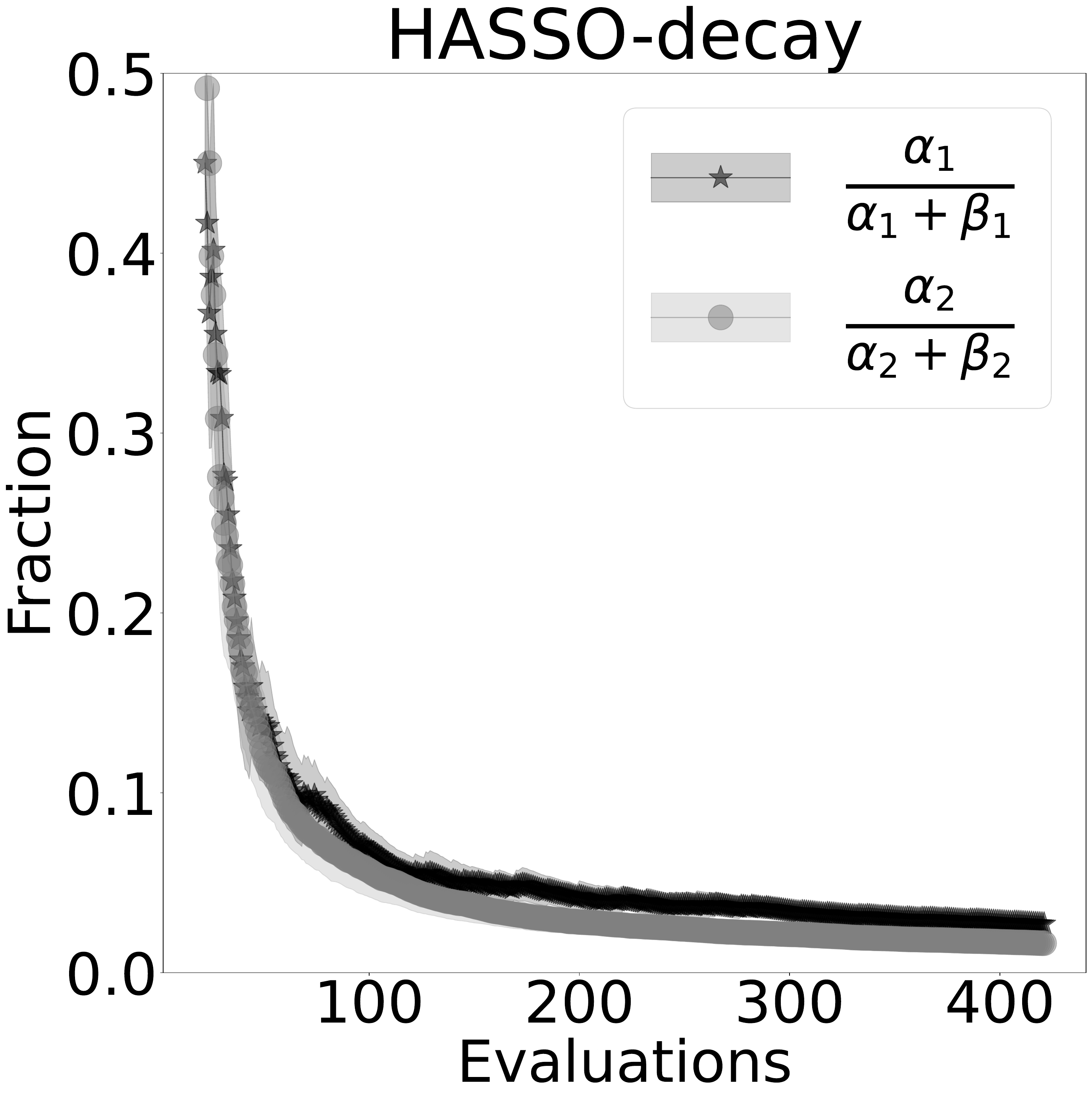}}
\subfigure[EI-Perm(10$d$)]{\label{hasso-perm}\includegraphics[width=0.24\linewidth]{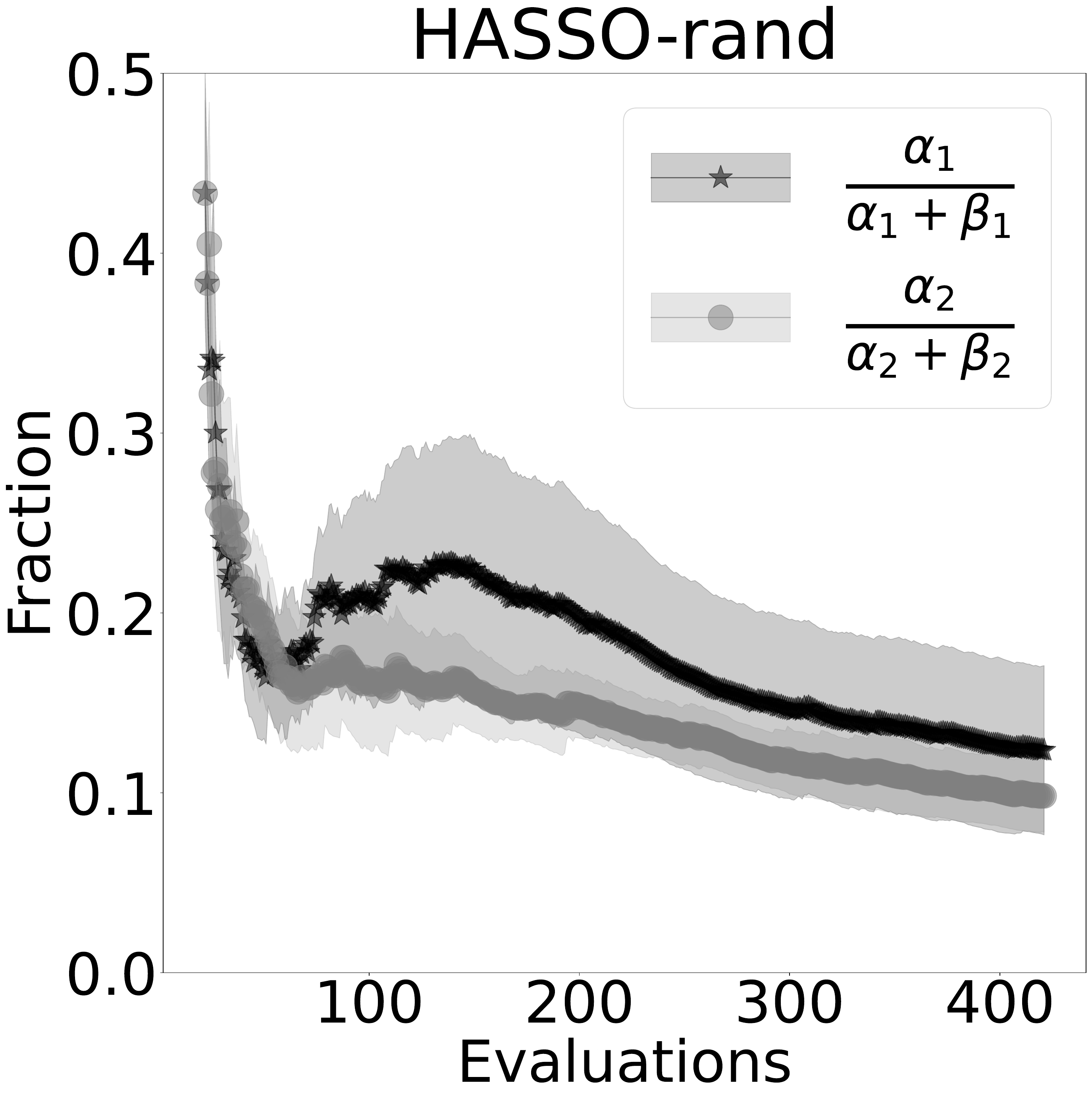}}
\subfigure[EI-Perm(10$d$)]{\label{decay-perm}\includegraphics[width=0.24\linewidth]{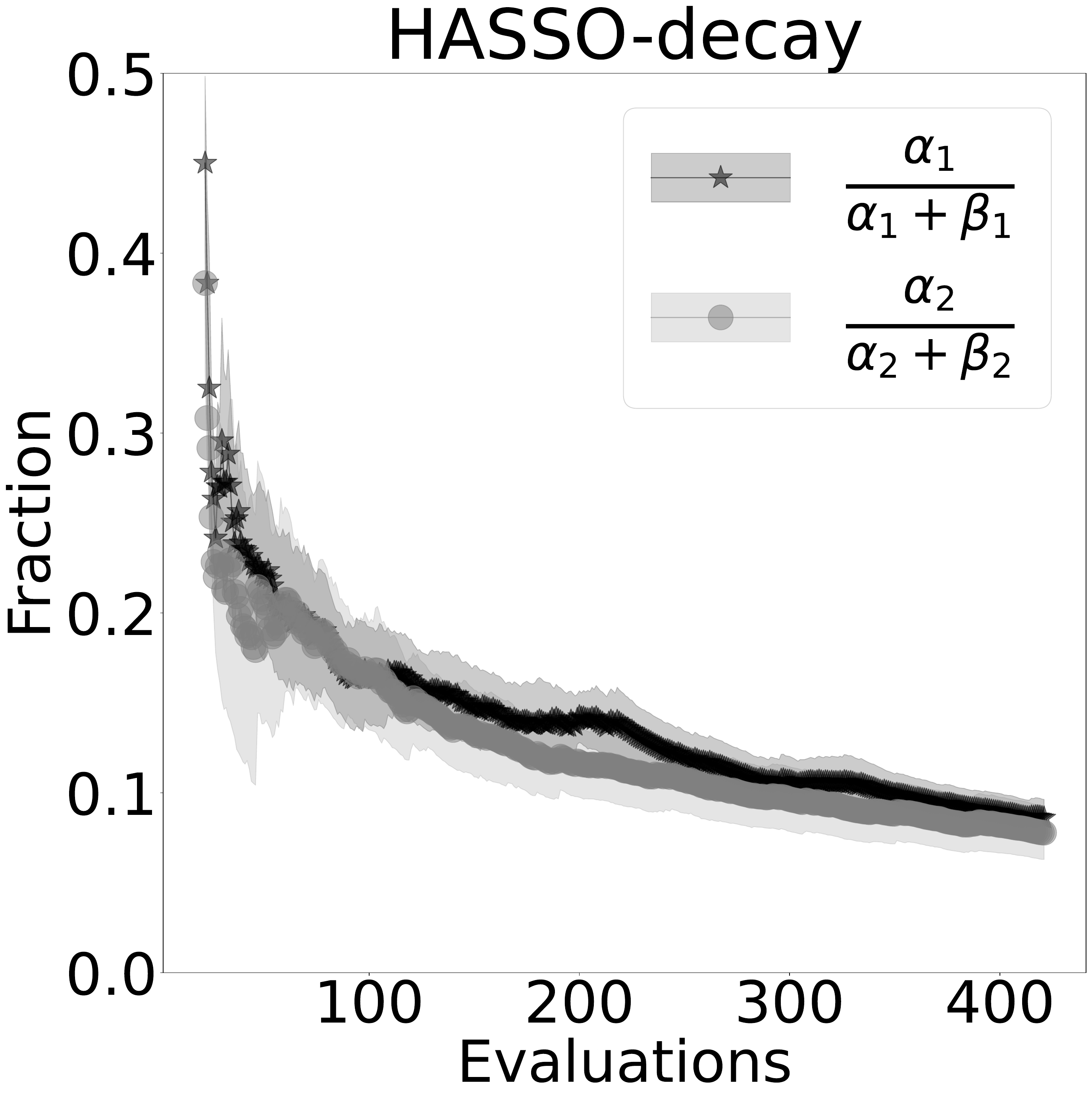}}
%\vspace{-3mm}
\caption{Impact of the update rule by comparing HASSO-rand with HASSO-decay.}
\label{exp:final2}
\end{figure}

In Figure~\ref{exp:final2}, we present the average $\frac{\alpha}{\alpha+\beta}$ fractions for the two hyperparameters in the test problems shown in Figure~\ref{exp:final}. The findings affirms that HASSO-rand outperforms HASSO-decay in terms of success rate (objective improvements) in both problems. This distinction is particularly pronounced for Rosenbrock. 
% where HASSO-rand significantly outperforms HASSO-decay. 
Furthermore, for HASSO-rand, the first hyperparameter (GP's lengthscale) demonstrates considerably more success compared to the second hyperparameter (radius). In contrast, HASSO-decay has a similar number of successes and failures for both hyperparameters. 
% Overall, the results suggest that HASSO-rand is a more effective adaptive search strategy than HASSO-decay for global optimization problems with diverse topological characteristics. 

% Figure~\ref{exp:final2} presents the average $\frac{\alpha}{\alpha+\beta}$ fractions for the two hyperparameters in the considered test problems in Figure~\ref{exp:final}. The results confirm that HASSO has obtained more success (objective improvements) than failure (no improvement) compared with HASSO-decay in both problems. This happens where it significantly outperforms HASSO-decay (Rosenbrock) and also where it performs similarly (Perm). Specifically, the number of successes obtained from modifying the first hyperparameter (GP's lengthscale) is considerably higher than the second hyperparameter (radius) for HASSO. However, the two hyperparameter changes have induced a similar number of successes and failures when using HASSO-decay. 
% \vspace{-3mm}

\begin{wrapfigure}{r}{0.38\textwidth}
\vspace{-3 mm}
  %\begin{center}
    \includegraphics[width=0.35\textwidth]{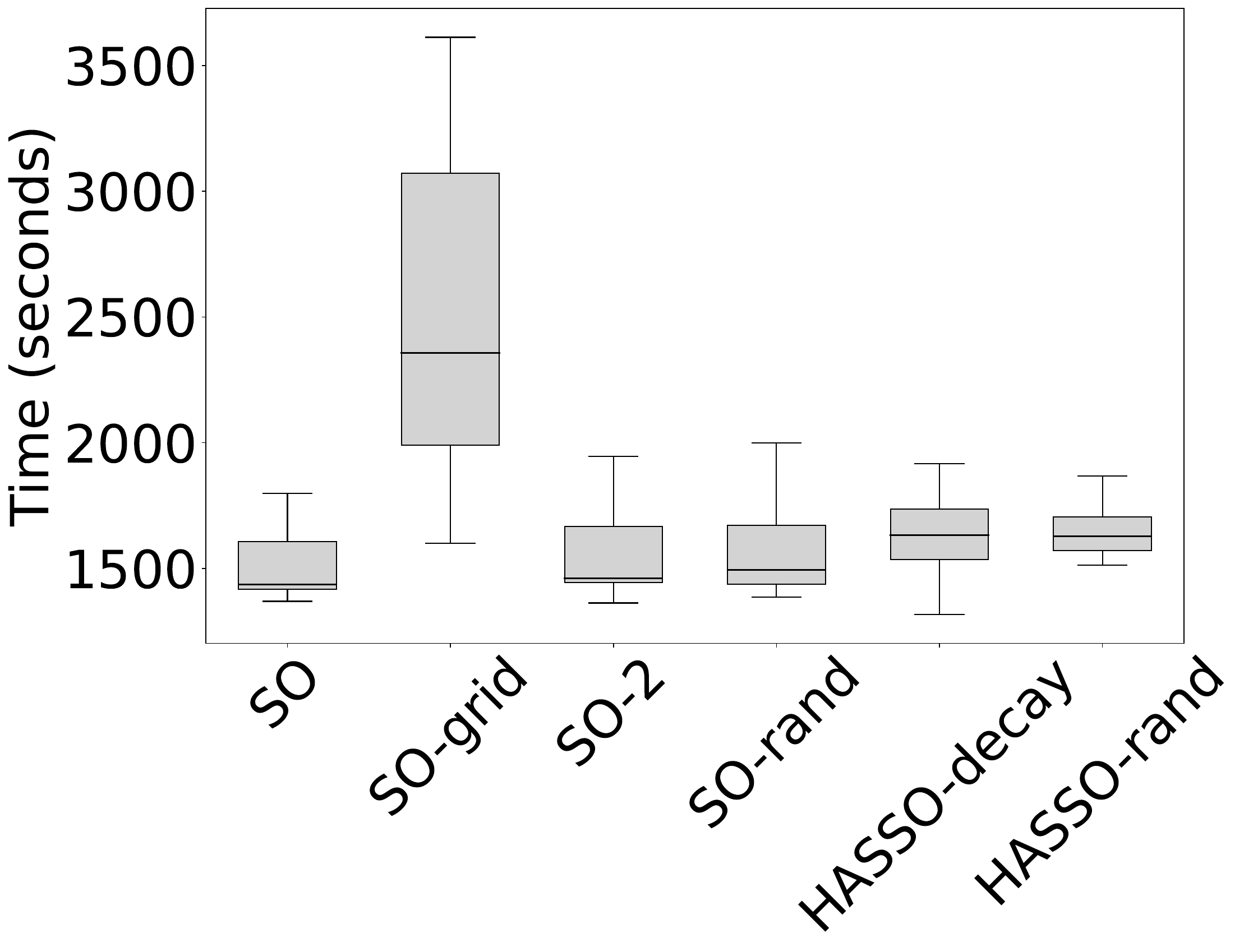}
  %\end{center}
  \vspace{-4mm}
  \caption{Computational time to complete one full run of 30$d$ test problems.}
  \label{exp:time}
% \vspace{-15 mm}
\end{wrapfigure}
% The boxplots in Figure~\ref{exp:time} presents a comparison of the computational time required to complete one full run (i.e., one repetition) of our considered baselines for high-dimensional (30$d$) global optimization test problems. The results indicate that incorporating HASSO into any regular SO algorithm results in only an insignificant increase in computational time when compared to the SO algorithm without HASSO (the SO baseline). Additionally, the computational time required for HASSO is insignificantly different from that of applying any random update (SO-rand) or rule (SO-2). However, it is worth noting that the use of a hierarchical grid search to identify the best hyperparameters can significantly reduce search efficiency.
In Figure~\ref{exp:time}, the boxplots present a comparison of the computational time required for a full run (i.e., one repetition) of our considered baselines for high-dimensional (30$d$) global optimization test problems. The results show that incorporating HASSO into any regular SO algorithm results in only a negligible increase in computational time compared to the SO algorithm without HASSO (the SO baseline). Additionally, the computational time required for HASSO is insignificantly different from that of applying any random update (SO-rand) or rule (SO-2). However, it is worth noting that using a hierarchical grid search to identify the best hyperparameters can significantly reduce search efficiency.

% \begin{figure}[!htb]
% \centering
% \includegraphics[width=0.3\linewidth]{Plots/time.pdf}
% \caption{Computational Time to Complete One Full Run of 30$d$ Test Problems}
% \label{exp:time}
% \end{figure}

%\nazanin{3 curve for hasso (UCB,EI,Wscore) using different test problems}

\section{CONCLUSION}\label{sec:conclusion}

% This study highlights that the performance of widely-used SO algorithms for solving expensive BBO problems is significantly impacted by multiple algorithmic hyperparameters. However, manual tuning of these hyperparameters can be challenging, thereby hindering the adoption of SO algorithms across various problems. To address this issue, we propose HASSO, which adaptively adjusts hyperparameter values based on the algorithm's outcome in each iteration of SO. The experimental results demonstrate the effectiveness of the proposed approach in improving the performance of various SO algorithms. HASSO identifies and modifies the most influential hyperparameters for a given problem and SO approach, reducing the need for manual tuning. Future directions should focus on expanding SO for parallel optimization settings and testing it with the consideration of more than two hyperparameters. Furthermore, it would be beneficial to investigate the performance of the proposed approach on real-world BBO problems.

This study sheds light on the crucial role played by hyperparameters in the performance of expensive black-box optimization algorithms. Manual tuning of these hyperparameters can be challenging and time-consuming, hindering the adoption of SO approaches across different problem domains. To address this issue, we proposed HASSO, an adaptive hyperparameter optimization framework that automatically adjusts hyperparameters based on the algorithm's outcome in each iteration of optimization. Our experimental results demonstrate that HASSO significantly improves the performance of various optimization algorithms while reducing the need for manual tuning.
Future research directions could explore the extension of HASSO to parallel optimization settings and investigate its performance in problems involving more than two hyperparameters. Additionally, testing the proposed approach on real-world optimization problems would provide valuable insights into its practical applicability. Overall, our study highlights the potential of adaptive hyperparameter optimization approaches like HASSO in enhancing the efficiency and effectiveness of black-box optimization algorithms.

% Reducing font size (to 9pt) for References & Author Biagraphies
\footnotesize
\section*{AUTHOR BIOGRAPHIES}
\noindent {\bf NAZANIN NEZAMI} is a Ph.D. Student in the Mechanical and Industrial Engineering Department at the University of Illinois Chicago (UIC). She obtained her M.S. degree in Industrial and Systems Engineering from the University of Minnesota Twin Cities prior to joining UIC. Her main research interests are Black-Box Optimization, Machine Learning (ML), and Fairness in ML. Her email address is \url{nnezam2@uic.edu}. \\

\noindent {\bf HADIS ANAHIDEH} is an Assistant Professor in the Mechanical and Industrial Engineering Department at the University of Illinois Chicago. She received her Ph.D. degree in Industrial Engineering from the University of Texas at Arlington. Her research objectives center around Black-box Optimization, Sequential Optimization, Active Learning, Statistical Learning, Explainable AI, and Algorithmic Fairness. Her email address is \url{hadis@uic.edu} and her homepage can be found at \url{https://mie.uic.edu/profiles/anahideh-hadis/}.

%\vspace{-5mm}
% Please don't exchange the bibliographystyle style
\bibliographystyle{wsc}
% AUTHOR: Include your bib file here
\bibliography{ref}

\end{document}